\documentclass[runningheads]{llncs}
\usepackage{graphicx}

\usepackage{graphicx}
\usepackage{amsmath}
\usepackage{amssymb}
\usepackage{booktabs}
\usepackage{adjustbox}
\usepackage{dirtytalk}
\usepackage[caption=false]{subfig}
\usepackage{algorithm}
\usepackage{algpseudocode}

\usepackage{hyperref}
\hypersetup{
    colorlinks=true,
    linkcolor=blue,
    citecolor=black,
}

\usepackage{xcolor}
\definecolor{darkpastelgreen}{rgb}{0.01, 0.80, 0.01}
\definecolor{seabornpurple}{rgb}{0.6312, 0.34, 0.86}

\usepackage{tikz}
\usepackage{comment}
\usepackage{amsmath,amssymb} 
\usepackage{color}

\newcommand{\etal}{\emph{et al}.}

\usepackage{times}
\usepackage{epsfig}
\usepackage{graphicx}
\usepackage{float}
\usepackage{wrapfig}
\usepackage{amsmath,amssymb} 
\usepackage{algorithm,algorithmicx,algpseudocode}
\usepackage{bm,xspace}
\usepackage{comment}
\usepackage{verbatim}
\usepackage{multirow}
\usepackage{balance}
\usepackage{url}
\usepackage{booktabs}
\usepackage{etoolbox,siunitx}
\usepackage{calc}
\usepackage{pifont,hologo}
\usepackage{nicefrac}

\definecolor{TableauOrange}{RGB}{242,142,44}
\definecolor{TableauGreen}{RGB}{89,161,79}

\usepackage[super]{nth}
\usepackage{nicefrac}
\robustify\bfseries

\usepackage{dsfont}

\def\tt{\mathbf{t}}

\DeclareMathSymbol{@}{\mathord}{letters}{"3B}

\def\latex/{\LaTeX}
\def\bibtex/{\hologo{BibTeX}}

\usepackage{authblk}
\makeatletter
\renewcommand\AB@affilsepx{, \protect\Affilfont}
\makeatother

\usepackage[accsupp]{axessibility}  

\begin{document}
\pagestyle{headings}
\mainmatter
\def\ECCVSubNumber{1039}  

\title{SALVe: Semantic Alignment Verification \\ for Floorplan Reconstruction from Sparse  Panoramas } 


\titlerunning{SALVe}
%

\author{John Lambert\inst{2} \makebox[0pt]{\thanks{Work completed during an internship at Zillow Group.}} \and
Yuguang Li\inst{1} \and
Ivaylo Boyadzhiev\inst{1} \and
Lambert Wixson \inst{1} \and
Manjunath Narayana \inst{1} \and
Will Hutchcroft \inst{1} \and
James Hays \inst{2} \and
Frank Dellaert \inst{2} \and
Sing Bing Kang \inst{1}} 

\authorrunning{J. Lambert et al.}
%
\institute{Zillow Group \and
Georgia Institute of Technology
}
\maketitle

\begin{abstract}

We propose a new system for   automatic 2D floorplan reconstruction that is enabled by {\em SALVe}, our novel pairwise learned alignment verifier. The inputs to our system are sparsely located 360$^\circ$ panoramas, whose semantic features (windows, doors, and openings) are inferred and used to hypothesize pairwise room adjacency or overlap. SALVe initializes a pose graph, which is subsequently optimized using GTSAM~\cite{Dellaert12_GTSAM}. Once the room poses are computed, room layouts are inferred using HorizonNet~\cite{Sun19cvpr_HorizonNet}, and the floorplan is constructed by stitching the most confident layout boundaries. We validate our system qualitatively and quantitatively as well as through ablation studies, showing that it outperforms state-of-the-art SfM systems in completeness by over 200\%, without sacrificing accuracy. Our results point to the significance of our work: poses of 81\% of panoramas are localized in the first 2 connected components (CCs), and 89\% in the first 3 CCs. Code and models are publicly available at \href{https://github.com/zillow/salve}{github.com/zillow/salve}. 



\end{abstract}

\section{Introduction}
\label{sec:intro}

Indoor geometry reconstruction enables a variety of applications that include virtual tours, architectural analysis, virtual staging, and autonomous navigation. There are solutions for image-based reconstruction based on inputs ranging from dense image capture to sparser capture using specialized imaging equipment (e.g., Matterport Pro2). For scalability of adoption, however, data bandwidth, equipment costs, and amount of labor must be considered. 


We reconstruct floorplans from sparsely captured $360^\circ$ panoramas, as provided by ZInD~\cite{Cruz21cvpr_ZillowIndoorDataset}. 
Currently, this problem is far from solved. Traditional Structure-from-Motion (SfM) \cite{Moulon16iwrrpr_OpenMVG,Gargallo16github_OpenSfM} suffers from very limited reconstruction completeness \cite{Cruz21cvpr_ZillowIndoorDataset,Shabani21iccv_ExtremeSfM}. 
Semantic SfM has been proposed \cite{Bao11cvpr_SemanticSfM,Cohen15iccv_StitchingDisconnected,Cohen16eccv_IndoorOutdoorAlignment}, but accuracy is still limited, typically requiring a human in the loop \cite{Cruz21cvpr_ZillowIndoorDataset}. 

\begin{figure}[t]
    \centering
    \includegraphics[width=0.97\columnwidth]{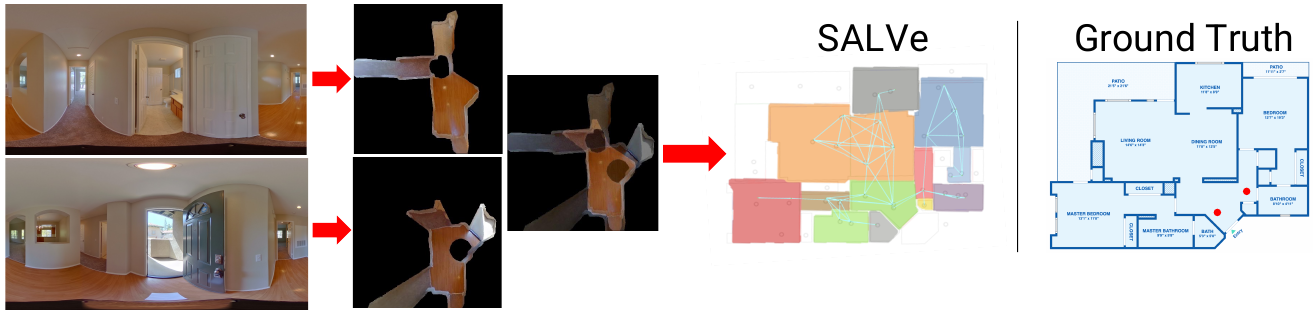}
    \caption{A challenging wide-baseline scenario where  traditional SfM systems that rely upon keypoint feature matches struggle, but where we succeed by exploiting semantic features such as windows, doors, and openings, or W/D/O). We infer layout and hypothesize plausible pairwise relative poses, which are then accepted or rejected, by feeding top-down aligned renderings into our learned {\em SALVe} verifier. Our global pose estimation has high completeness, leading to dramatic improvements in floorplan reconstruction (indicated by colored regions) vs. state-of-the-art systems such as OpenMVG \cite{Moulon16iwrrpr_OpenMVG} and OpenSfM \cite{Gargallo16github_OpenSfM}. For this hallway/entryway pano pair, SALVe easily validates a relative pose that was generated by grounding on a hallway opening feature.}
    \label{fig:hero}
\end{figure}
Indoor floorplan reconstruction from unordered panoramas is a \emph{discrete} instance of the wide-baseline SfM problem.  Unlike traditional SfM, which is associated with a continuous estimation problem, for indoor residential floorplan reconstruction, discrete room pieces must align at specific junction points (such as doors and walls), similar to solving a jigsaw puzzle \cite{Lin19iccv_FloorplanJigsaw}.
We show how objects with repetitive structure, such as windows and doors, can be used to hypothesize room adjacency or overlap. Each hypothesis, i.e. a matched semantic element, provides a relative 2D room pose.
The main innovation of our work is {\em SALVe}, a learned pairwise room alignment verifier. Given a room pair alignment hypothesis, {\em SALVe} uses the bird's eye view (BEV) of floors and ceilings to predict the likelihood score of adjacency or overlap. Our use of a discrete combinatorial proposal step, followed by a learned deep verifier, is akin to recent trends in language models, for tasks requiring multi-step reasoning \cite{Cobbe2021TrainingVT,Shen21emnlp_GenerateAndRank}, as \say{\emph{Verifiers benefit both from their inherent optionality and from verification being a simpler task than generation in general.}} \cite{Cobbe2021TrainingVT,Lambert21neurips_TrustButVerifyHDMapChangeDetection}.

Once the relative poses are computed and verified, we perform global pose graph optimization using GTSAM~\cite{Dellaert12_GTSAM}.
Using the estimated poses and room layouts generated using HorizonNet~\cite{Sun19cvpr_HorizonNet}, we construct the floorplan by stitching these layouts.



Our contributions are:
\setlength{\parskip}{0pt}
\begin{itemize}
\setlength{\itemsep}{0pt}
\setlength{\parskip}{0pt}
\setlength{\parsep}{0pt}
    \item To our knowledge, the first system for creating floorplans from unaligned panoramas with small to extremely wide baselines. These baselines can be so large that traditional SfM techniques fail.
    \item SALVe, a novel learning-based approach for validating discrete pairwise alignment proposals between panoramas in polynomial time.
    \item We show how our network verifies measurements with a high enough signal-to-noise ratio to directly apply global aggregation and optimization techniques.
\end{itemize}

\section{Related Work}


We briefly review approaches in floorplan reconstruction, SfM, and pose estimation under extreme baselines. 
While single-room layout estimation and depth estimation are also relevant, we do not claim novelty in these areas. Good surveys of such methods can be found in \cite{Pintore20cgf_SotaReconstructionIndoorSurvey} and \cite{Albanis2021Pano3DAH}.


\noindent \textbf{Floorplan Reconstruction.} 
Early systems require a human in the loop~\cite{Debevec96ccgit_ImageBasedRendering,Farin07acmicm_FloorplanPanoramicImages}. One notable manual approach is that of Farin \etal~\cite{Farin07acmicm_FloorplanPanoramicImages}, which uses sparsely located 360$^\circ$ panoramas for joint floorplan and camera pose estimation.

For more automated solutions, SfM is used on densely captured perspective images~\cite{Furukawa09iccv_ReconstructingBuildingInteriors} or 360$^\circ$ panoramas~\cite{Cabral14cvpr_FloorplanReconstructionImages}. Both use SfM and MVS output to formulate graph optimization problems on a regular grid, through either graph cuts~\cite{Furukawa09iccv_ReconstructingBuildingInteriors} or shortest-path problems~\cite{Cabral14cvpr_FloorplanReconstructionImages}, from which a rough 2D floorplan can be extracted. 
For sparser image inputs, semantic information such as floors, ceilings, and walls are used as additional cues~\cite{Pintore18cvm_3dFloorplanOverlappingSphericalImages}. Pintore \etal~\cite{Pintore19cgf_AutomaticFloorplan} cluster panoramas by room using photo-consistency at the central horizon line and plane sweeping with superpixel object masks to model clutter and floorplans in 3D.
There are also methods on floorplan reconstruction from known camera poses~\cite{Liu18eccv_FloorNet3dScans,Chen19iccv_FloorSPInverseCAD,Lin19iccv_FloorplanJigsaw,Stekovic21iccv_MonteFloor,Purushwalkam21iccv_AudioVisualFloorplan} or RGBD data~\cite{Liu18eccv_FloorNet3dScans,Chen19iccv_FloorSPInverseCAD,Lin19iccv_FloorplanJigsaw,Stekovic21iccv_MonteFloor,Purushwalkam21iccv_AudioVisualFloorplan,Okorn10_TowardAutomatedFloorPlans,Kim12icra_InterativeFloorPlans,Fang21jprs_IndoorSceneReconstructionTwoLevel,Fang21jprs_PointCloudFloorPlan}.

\noindent \textbf{Structure from Motion (SfM).} Much work has been done on SfM, and we refer readers to surveys such as \cite{Ozyesil2017}. Recently, deep learning with graph-based attention~\cite{Sarlin20cvpr_SuperGlue} or transformers~\cite{Sun21cvpr_LoFTR} for deep, differentiable key point matching has been exploited to learn and match features from data. These ``deep front-ends" offer a promise of less noisy input to back-end optimization~\cite{Sarlin20cvpr_SuperGlue}. Our system can be viewed as a deep verifier network (a deep front-end) that feeds measurements to global SfM \cite{Moulon13iccv_GlobalFusionSfM,Sweeney15acmicm_TheiaSfM}; however, instead of requiring complex outlier rejection schemes typical of global SfM \cite{Enqvist11iccvw_NonsequentialSfM,Moulon13iccv_GlobalFusionSfM,Moulon16iwrrpr_OpenMVG,Zach10cvpr_LoopConstraints,Sweeney15acmicm_TheiaSfM,Wilson14eccv_1DSfM}, we show that outlier rejection can simply be based on predicted scores. 

Semantic information has been used to overcome the limitation of keypoint matching for large baselines or scenes with little detail or repetitive textures~\cite{Bao11cvpr_SemanticSfM,Choudhary14iros_ObjectDiscoverySLAM}. 
Cohen \etal~\cite{Cohen15iccv_StitchingDisconnected}  first introduced a combinatorial approach for 3D model registration by aligning semantic objects such as windows \cite{Cohen16eccv_IndoorOutdoorAlignment}. More recent work \cite{Cruz21cvpr_ZillowIndoorDataset,Shabani21iccv_ExtremeSfM} exploits this same idea to assemble floorplans from room layouts.





\noindent \textbf{Extreme Pose Estimation.} 
This refers to computing relative pose with little to no visual overlap. On localizing RGBD images, Yang \etal~\cite{Yang19cvpr_ExtremeRelativePoseRGBD,Yang20cvpr_ExtremeRelativePoseNetwork} demonstrate scan completion to a $360^\circ$ image, followed by feature-based registration  can  be  useful. Chen \etal~\cite{Chen21cvpr_WideBaselineDirectionNet}  introduce  DirectionNet to estimate a distribution of relative poses in 5 DOF space, i.e., when scale is unknown. SparsePlanes \cite{Jin21iccv_SparsePlanes} uses planar surface estimation from perspective views within a single room for relative pose estimation. Other CNN-based approaches on perspective image re-localization include  \cite{Laskar17iccv_RelativePoseCNN,Balntas18eccv_Relocnet,ding2019camnet}.

In concurrent work, Shabani \etal~\cite{Shabani21iccv_ExtremeSfM} use semantic information to generate global pose hypotheses by synthesizing Manhattan-only floorplans. The hypotheses are then scored by ConvMPN \cite{Zhang20cvpr_ConvMPN} and used to produce plausible room layout arrangements along with camera poses. They assume each panorama is captured in separate but connected rooms.  Another key difference from our work is that their learning-based verifier is trained to evaluate the \emph{final floorplan arrangements}, after using heuristics to enumerate many possible solutions. This is \emph{exponential} in the number of input panoramas. Their approach is expected to produce several layout arrangements. In contrast, {\em SALVe} matches semantic elements between pairs of panoramas in polynomial time. Our model is then trained to verify the individual pairwise arrangements, allowing our approach to be substituted as a front-end in any pose-graph optimization and producing a single reconstruction with higher reliability.

\section{System Overview}

We address the problem of global pose estimation of sparsely located panoramas, for the purpose of floorplan reconstruction.   Formally, we define the global pose estimation problem as, given an unordered collection of $n$ panoramas  $\{\mathbf{I}_i\}$, determine poses $\{ {}^w \mathbf{T}_i \}_{i=1}^n \in SE(2)$ of each panorama in global coordinate frame $w$. 
Similar to \cite{Purushwalkam21iccv_AudioVisualFloorplan}, we define the floorplan reconstruction problem as generating a \emph{raster} (1) floor occupancy and (2) per-room masks.



\begin{figure*}[t]
    \centering
    \includegraphics[width=1\columnwidth]{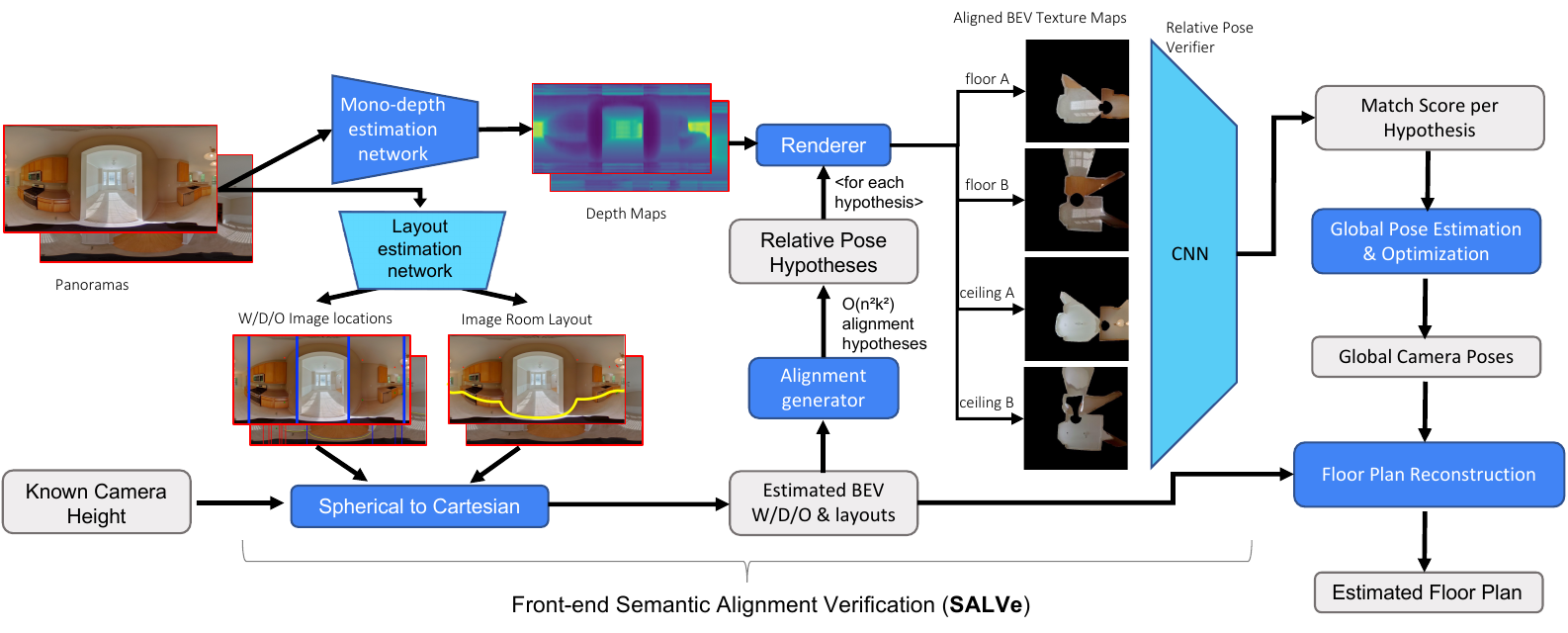}
    \vspace{-8mm}
    \caption{Overview of our floorplan reconstruction system.
    ``BEV" = ``bird's eye view". Blue boxes are processing components, gray boxes are data.  Trapezoids denote components based on deep networks; lighter blue networks are trained by us. `Image Room Layout' represents the image coordinates of the floor-wall boundary (at each panorama column). $n$ is the number of panoramas and $k$ is the average number of detected windows/doors/openings per panorama. We show rendered floor and ceiling texture maps for a consistently-aligned pair of panoramas.  
    }
    \label{fig:system-block-diagram}
    \vspace{-4mm}
\end{figure*}

Global pose estimation inherently relies on methods that build up global information from local signals.  In our work, these local signals are estimated relative poses between pairs of panoramas.  Our system for generating the floorplan from sparsely located panoramas is shown in Figure~\ref{fig:system-block-diagram}. The system consists of a front-end designed to hypothesize and compute relative pairwise poses, and a back-end designed to optimize global poses using these measurements. 

The front-end ({\em SALVe}, or Semantic Alignment Verifier) first generates hypotheses of relative pose between the input pair of panoramas using their estimated room layout and detected semantic objects (specifically windows, doors, and openings, or W/D/O).\footnote{Openings are constructs that divide a large room into multiple parts~\cite{Cruz21cvpr_ZillowIndoorDataset}.} A hypothesis consists of pairing the same type of object across the two panoramas. Each pair of hypothesized corresponding W/D/O detections generates two relative pose hypotheses, by solving for the 2D translation that aligns their centers (on the ground plane), and the two possible rotation angles \(\alpha, 180^\circ+\alpha \) that align their extents. Each pairing allows us to compute the relative SE(2) pose.

A main novelty in this paper is how we test whether a hypothesis is plausible with {\em SALVe}.  For a hypothesized relative pose, the system renders bird's-eye views of the floor and ceiling for both panoramas in the same BEV coordinate system, which produces overlapped top-down renderings. The rendering is computed with per-panorama depth distribution estimation using HoHoNet \cite{Sun21cvpr_HoHoNet}. Then we use a deep CNN with a ResNet \cite{He16cvpr_ResNet} backbone to generate a likelihood score that the overlapped images are a plausible match. Implausible matches are discarded, and from the remaining plausible matches we construct a pose graph.  The back-end then globally optimizes the constructed pose graph using GTSAM~\cite{Dellaert12_GTSAM}.
Finally, floorplans are created by clustering the panoramas by room, extracting the most confident room layout given predicted panorama poses, and finally stitching these room layouts.


\section{Approach}
In this section, we detail the steps taken to generate a 2D floorplan from sparsely distributed 360$^\circ$ panoramas. The first step is to generate alignment hypotheses between pairs of panoramas.

\subsection{Assumptions}

We assume the inputs are a set of unordered $360^\circ$ panoramas, captured from an indoor space. 
The images cover the entire space and the connecting doors between different rooms. 
Neighboring images may or may not have visual overlap.
We assume the panoramas are in  equirectangular form, i.e., their fields of view are $360^\circ$ (horizontal) and $180^\circ$ (vertical). The camera is assumed to be of known height and fixed orientation parallel to the floor\footnote{We achieve this orientation assumption via pre-processing that straightens the panoramas using vanishing points~\cite{Zhang14eccv_PanoContext}.}, so pose is estimated in a 2D bird's-eye view (BEV) coordinate system. 

\subsection{Generating Alignment Hypotheses} \label{sec:generate-alignment-hypotheses}

Since our floorplan is 2D, alignment between pairs of panoramas has 3 DOFs (horizontal position and rotation). Scale is not a free parameter, assuming known, fixed camera height and a single floor plane (see \cite{Aly12wacv_StreetviewIndoors} or our Appendix for a derivation). To handle wide baselines, we use semantic objects (windows, doors, and openings, or W/D/O) to generate alignment hypotheses. While this is similar to the W/D/O-based room merge process in \cite{Cruz21cvpr_ZillowIndoorDataset}, we additionally make use of estimated room layout. Each room layout is estimated using a modified HorizonNet model \cite{Sun19cvpr_HorizonNet}; it is trained with partial room shape geometry to predict both the floor-wall boundary with an uncertainty score and locations of W/D/O.


Each alignment hypothesis is generated with the assumption that W/D/O being aligned are in either the same room or different rooms. The outward surface normals of W/D/O are either in the same or opposite directions; we assume a window can only be aligned in the direction of its interior normal, while a door or opening could be aligned in either direction. 
The hypothesis for rotation is refined using dominant axes of the two predicted room layouts.


Exhaustively listing pairs of W/D/O can produce many hypotheses for alignment verification. We halve the combinatorial complexity by ensuring that each pair of matched W/D/O have widths with a ratio within $[0.65, 1.0]$, i.e. a door that is 2 units wide cannot match to a door that is 1 units wide. Once the alignment hypotheses are found, they need to be verified.


\vspace{-2mm}
\subsection{SALVe: Semantic Alignment Verifier}
\vspace{-0.5mm}
While domain knowledge of indoor space such as room intersections and loop closure can be helpful in constructing the floorplan \cite{Cruz21cvpr_ZillowIndoorDataset}, visual cues can also be used to verify pairwise panorama overlap  \cite{Dellaert1999cvpr_CondensationAlgorithmLocalization}. We use bird's eye views (BEVs, which are orthographic) of the floor and ceiling as visual cues for alignment verification. Given the significant variation in lighting and image quality across panoramas, traditional photometric matching techniques may not be very effective. Instead, we train a model to implicitly verify spatial overlap based on these aligned texture signals.

\begin{figure*}[t]
    \centering
    \includegraphics[trim=0 40 0 0,width=0.8\columnwidth]{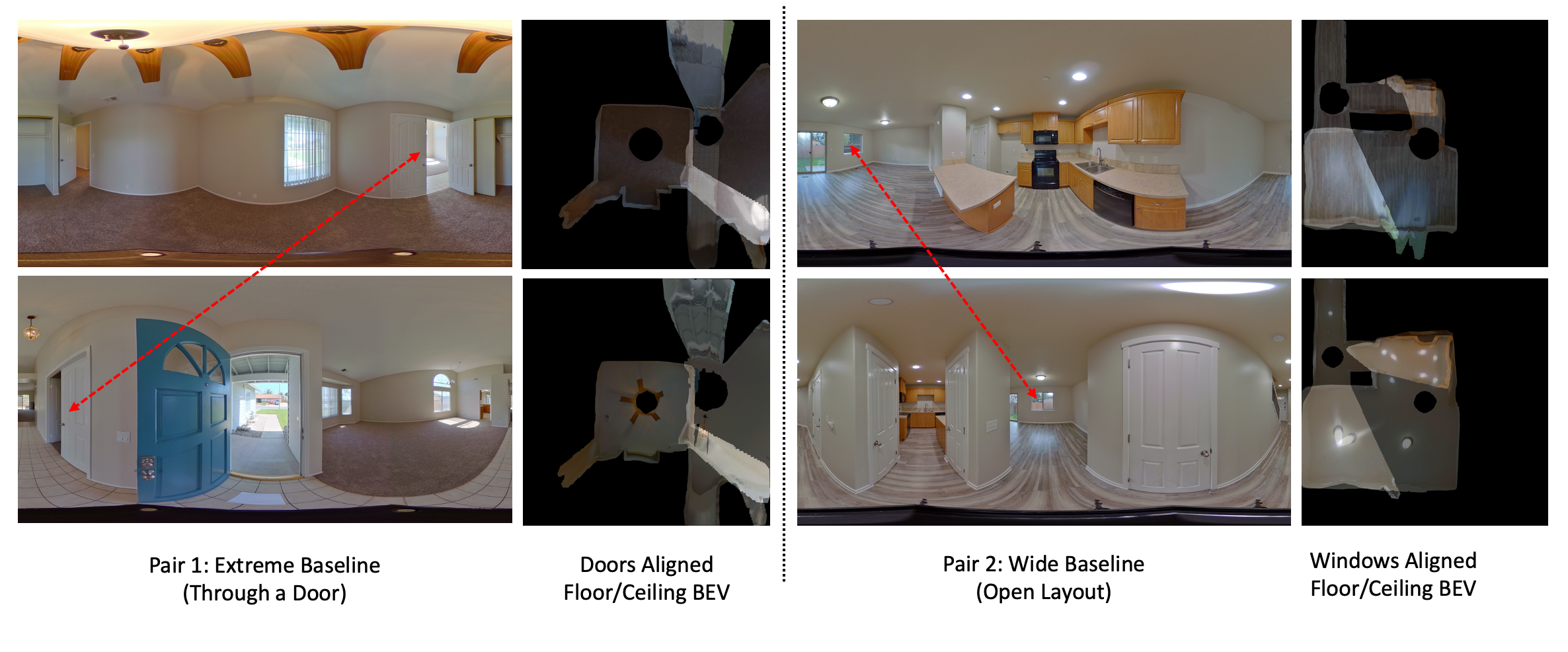}
    \caption{Generating training samples. Orthographic BEVs of given panoramas, after semantic alignment proposal. Red arrows indicate the W/D/O, used to generate the pose proposals.  \emph{\textbf{Column 1:}} Example of extreme baseline pair. \emph{\textbf{Column 2:}} overlaid floor \textbf{(top)} and ceiling \textbf{(bottom)}. \emph{\textbf{Column 3:}} Example of a wide baseline pair. \emph{\textbf{Column 4:}} overlaid floor \textbf{(top)} and ceiling \textbf{(bottom)}.}
    \label{fig:orthographic-samples}
\end{figure*}

We extract depth using HoHoNet\cite{Sun21cvpr_HoHoNet}, which is used to render the BEVs. Example views can be found in Figure~\ref{fig:orthographic-samples}.
Given an alignment hypothesis, we map the BEVs of the floor and ceiling for both panoramas to a common image coordinate system. The four stacked views are then fed into our deep-learning based pairwise alignment verification model to classify 2-view alignment. Given $n$ panoramas, each with $k$ W/D/O, $\mathcal{O}(n^2k^2)$ alignments are possible and thus need to be verified.

{\em SALVe} uses a ResNet \cite{He16cvpr_ResNet} ConvNet architecture as the backbone for verification \cite{Lambert21neurips_TrustButVerifyHDMapChangeDetection}. Its input is a stack of 4 aligned views (2 from each panorama), with a total of 12 channels. It is trained with softmax-cross entropy over 2 classes, representing the ``mismatch" and ``match" classes. 
We generate these classes by measuring the deviation of generated relative poses (alignments from window-window, opening-opening, or door-door pairs) against the ground truth poses. Those below a certain amount of deviation are considered ``matches", and all others are considered ``mismatches".

\subsection{Global Pose Estimation and Optimization}\label{sec:global-pose-estimation-optimization}

{\em SALVe} is used to generate a set of pairwise alignments, which are used to construct a pose graph; its nodes are panoramas and edges are estimated relative poses. The pose graph has an edge between any two panoramas $\mathbf{I}_{i_1}$ and $\mathbf{I}_{i_2}$ where pairing a detection $\mathbf{d}_{k_1}^{i_1}$ with detection $\mathbf{d}_{k_2}^{i_2}$ yields a plausible (according to SALVe) alignment. A detection may participate in multiple edges e.g., pairing ($\mathbf{d}_{k_1}^{i_1}, \mathbf{d}_{k_2}^{i_2}$) may add an edge between $i_1$ and $i_2$, and pairing ($\mathbf{d}_{k_1}^{i_1}, \mathbf{d}_{k_3}^{i_3}$) may add an edge between panos $i_1$ and $i_3$. Although conflicting relative pose hypotheses are possible, in practice SALVe is a sufficiently accurate verifier that they are quite rare. 

When multiple disjoint graphs result, we only consider the largest connected component. We experiment with two algorithms for global localization: spanning tree pose aggregation and pose graph optimization (PGO) with a robust noise model, detailed in the Appendix.



\subsection{Floorplan Reconstruction}
\label{sec:floorplan-reconstr-approach}
\begin{figure}[t]
    \centering
    \includegraphics[trim=0 0 0 0,width=0.6\columnwidth]{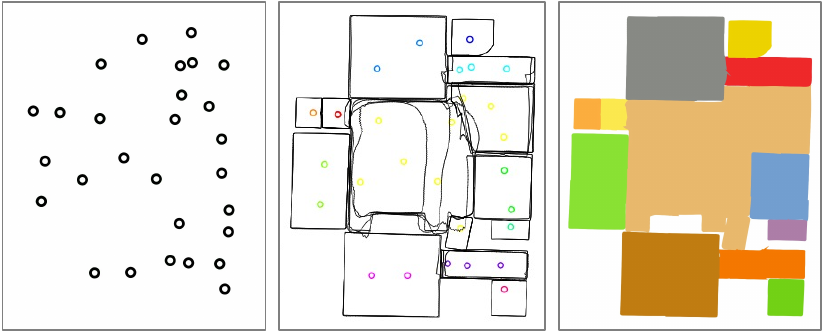}
    \caption{An example of different stages of floorplan reconstruction: \textbf{\emph{Left}}: Estimated positions of panorama centers.  \textbf{\emph{Center}}: Grouped panoramas with estimated dense room layouts.   Panorama centers with the same color are part of the same group.  Notice that each open space is grouped together.  Distinct groups correspond largely to physical rooms separated by doors.  \textbf{\emph{Right}}: The final floorplan after highest-confidence contour extraction is applied to each group.   Each contour is filled with a unique color.}
    \label{fig:floor-plan-stitching}
\end{figure}

Figure~\ref{fig:floor-plan-stitching} shows the progression of floorplan reconstruction, from estimated panorama poses and room layouts to the output. There are three steps: panorama room grouping, highest confidence room contour extraction, and floorplan stitching. To refine a room layout, we first identify all the panoramas within that room; this is done using 2D IoU. Since each panorama has its own layout with local shape confidence (Section \ref{sec:generate-alignment-hypotheses}) within a room, we extract a single global layout by searching for the most confident contour points. The search is done by raycasting from panorama centers and voting for the most confident contour point along each ray. The final floorplan is found by taking the union of (stitching) all room layouts. Details are in the Appendix.

\section{Experimental Results}

In this section, we explain why we use ZInD~\cite{Cruz21cvpr_ZillowIndoorDataset}, provide implementation details, and describe our metrics before showing results for different global pose estimation techniques. We also describe ablation studies that show how different types of inputs affect the results.

\subsection{Use of ZInD~\cite{Cruz21cvpr_ZillowIndoorDataset}} 

In order to evaluate every part of our approach, as well as the entire system, we use the recently released Zillow Indoor Dataset (ZInD)~\cite{Cruz21cvpr_ZillowIndoorDataset}. ZInD has all the required components: (1) \emph{large scale} with 67,448 panoramas taken in 1,575 real
homes; (2) \emph{multiple localized panoramas per-room} with $42$ panoramas over $15$ rooms per-home on average; (3) \emph{layout and W/D/O} annotations including complex, non-Manhattan layouts and (4) \emph{2D floor-plans} with $1.8$ number of floors per-home on average. We use the official train, val, and test splits that contain $1260$, $157$, and $158$ homes, and $2168$, $278$, $291$ floors respectively. We acknowledge that in ZInD most rooms are unfurnished, but this is a frequent scenario in the domain of real estate floor plan reconstruction. While there are other real~\cite{Matterport3D,Zou21ijcv_MatterportLayout,Shabani21iccv_ExtremeSfM} and synthetic~\cite{Zheng20eccv_Structured3d,Song17cvpr_SemanticSceneCompletion} indoor datasets, none of them have all the required components. Structured3D \cite{Zheng20eccv_Structured3d} is a synthetic dataset with only one panorama per room and doors in almost all rooms are closed (uncommon in real estate capture scenarios); these factors result in a significant change of modality. 

\subsection{Implementation Details}
\label{sec:impl-details-main-paper}
\noindent \textbf{Layout and W/D/O estimation.} We use a modified version of HorizonNet~\cite{Sun19cvpr_HorizonNet}, trained to jointly predict room layout as well as 1D extents of W/D/O. We trained the joint model on ZInD and make the predictions publicly available. 

\noindent \textbf{Verifier supervision.} We consider a pair-wise alignment to be a ``match" if ground truth relative pose $(x, y, \theta) \in SE(2)$  and generated relative pose $(\hat{x}, \hat{y}, \hat{\theta}) \in SE(2)$ differ by less than $7^\circ$ ($\theta$) for doors and windows, and less than $9^\circ$ for openings. A larger threshold is used for openings because there is more variation in their endpoints. We also require that $\big\| [x,y]^\top - [\hat{x},\hat{y}]^\top  \big\|_{\infty} < 0.35$ in normalized room coordinates (i.e., when camera height is scaled to 1).

\noindent \textbf{Texture mapping, verifier data augmentation and verifier training.} Details are provided in the Appendix.

\subsection{Evaluation Metrics}
In order to evaluate our entire system, we measure increasing subsets of components. 
\noindent \textbf{Layout estimation and W/D/O detection accuracy}. To evaluate the quality of the layout estimation, we report 2D IoU between the predicted and ground truth room layouts per panorama. Because we project 1D W/D/O on the predicted layout, we use 1D IoU to measure the accuracy of those semantic elements, with F1 score evaluated at a true positive 1D IoU threshold of 70\%.\\ 
\noindent \textbf{Relative pose classification accuracy}. We report intermediate system metrics that measure the model's accuracy at discerning between correct and inaccurate alignments. We use mean accuracy over two classes, as well as precision, recall, and F1 score.\\
\noindent \textbf{Global pose estimation accuracy and completeness}. 
We first align an estimated pose graph $\{ \hat{\textbf{T}}_i  \}_{i=1}^M$ to a ground truth pose graph $\{ \textbf{T}_i  \}_{i=1}^N$ where $\textbf{T}_i \in SE(2) \hspace{2mm} \forall i \in 1,\dots,N$, by estimating a $\mathrm{Sim}(2)$ transformation between them, where $M \leq N$, since not all poses may be estimated. To reduce the influence of outliers for mostly-correct global pose estimates, we perform pose graph alignment in a RANSAC loop using \cite{Baid23arxiv_GTSFM}, with a randomly selected subset ($\nicefrac{2}{3}$ of the $M$ estimated poses) used to fit each alignment hypothesis, over 1000 hypotheses. We then measure the distance between the predicted and true $i$'th camera location $\|t_i - \hat{t}_i \|_2$, and difference between true and predicted $i$'th camera orientation $|\theta_i - \hat{\theta}_i|$. Completeness is essential to floorplan reconstruction, so we also report the percent of panoramas localized in the largest connected component.\\
\noindent \textbf{Floorplan reconstruction accuracy and completeness.} We measure the 2D IoU between a rasterized binary occupancy map of the ground truth and the predicted floorplans. This metric measures the quality of our end-to-end system, as it encapsulates the \emph{accuracy} of our \emph{pair-wise relative pose proposal} in combination with the \emph{accuracy} and \emph{completeness} of the global pose estimation and the fusion of the room layouts (see Appendix for more details).






\begin{figure}[t]
    \centering
    \includegraphics[width=0.3\columnwidth]{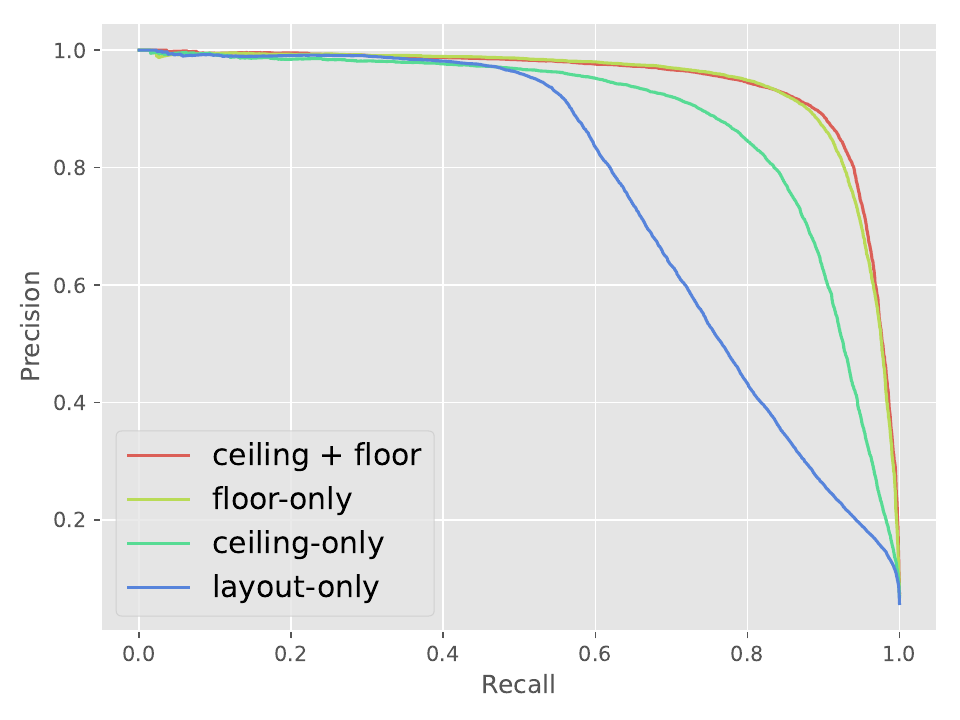}
    \includegraphics[width=0.3\columnwidth]{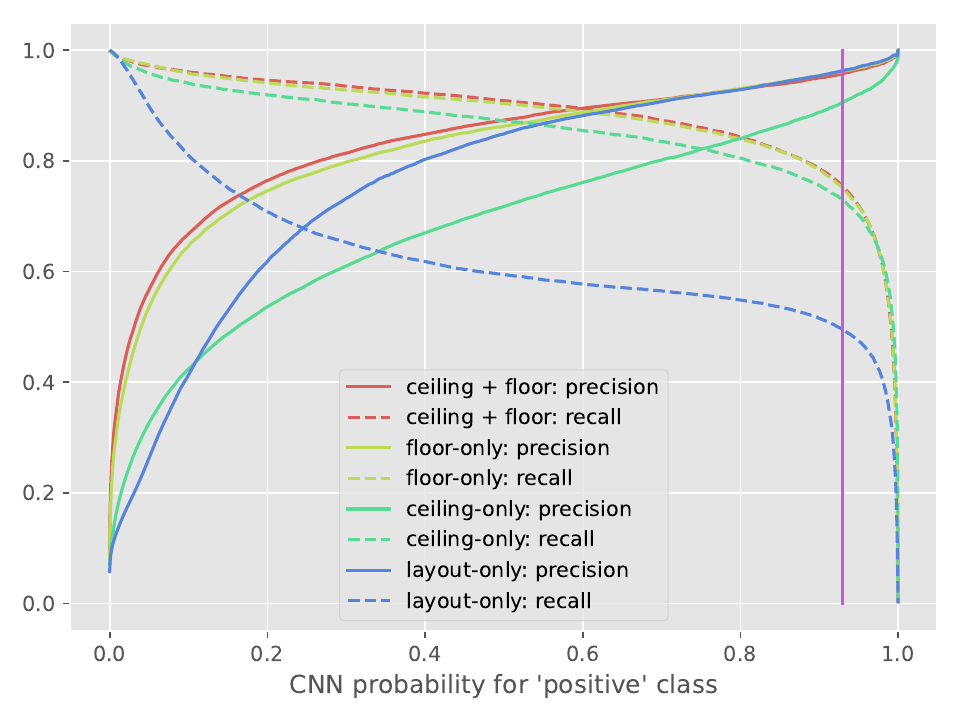}
        \includegraphics[trim=20 15 0 0,width=0.30\columnwidth]{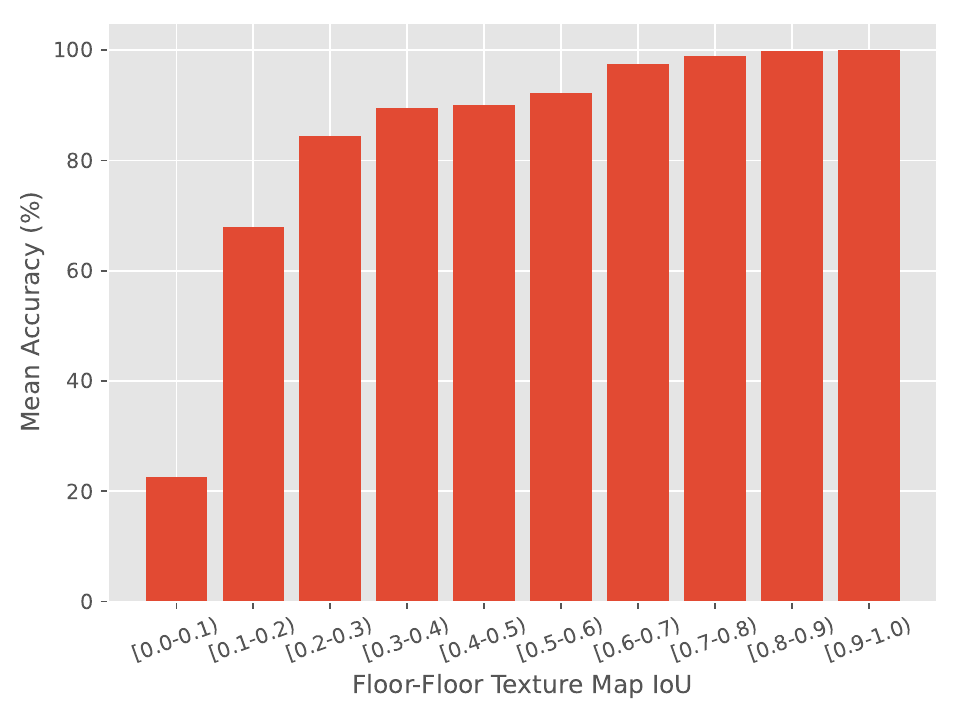}
    \caption{Precision-recall analysis of SALVe. \emph{\textbf{Left}}: curve for SALVe under different inputs (`layout-only' refers to a model with access only to estimated room geometry, but no floor or ceiling texture). \emph{\textbf{Center}}: Comparison of confidence thresholds versus their effect on precision and recall. The \textcolor{seabornpurple}{purple} line indicates our operating point (93\% confidence).
    \emph{\textbf{Right}}: Classification accuracy vs. visual overlap for the GT \textbf{positive} class only from SE(2) alignments generated from predicted W/D/O's. Small visual overlap often corresponds to ``extreme'' baselines.}
    \label{fig:binary-classification-pr-curve}
\end{figure}

\subsection{Layout and W/D/O Estimation Accuracy}
\label{sec:layout-wdo-estimation-acc-main-paper}
The layout estimation module used in the system yields an average of 85\% IoU with ground truth shape. 
W/D/O detection is accurate; at a 70\% 1D IoU threshold, we correctly identify W/D/O with F1 scores of 0.91, 0.89, and 0.67, respectively. Among categories, our model is the least accurate in predicting openings. As discussed in \cite{Cruz21cvpr_ZillowIndoorDataset}, there are issues with annotator error and possibly ambiguous tagging of rooms in open spaces that cover different room types, making locations of openings less clear. We speculate that these contribute to the errors, especially for openings. In the Appendix, we provide qualitative examples of the various types of failure modes of the model.




\subsection{Relative Pose Classification}
\label{sec:relative-pose-classification-results-main-paper}
We first measure the performance of the SALVe ``front-end''. These trained models achieve 92-95\% accuracy on the test split (see Appendix). We show that a larger capacity model than ResNet-50 (i.e. ResNet-152) further improves performance. We also note that the accuracy is limited by noisily-generated `ground truth'. We train on 587 number of tours from ZInD, and use the official train/val/test splits.

In Figure \ref{fig:binary-classification-pr-curve}, we show a PR curve, indicating the precision of the model at different recall thresholds. We choose a 93\% confidence threshold as our operating point, as it maximizes precision just before a precipitous drop in recall.

\noindent \textbf{How does the amount of visual overlap affect relative pose classification accuracy?} More overlap yields higher accuracy for the ground truth positive class, but lower accuracy for the ground truth negative class. In Figure \ref{fig:binary-classification-pr-curve}, we analyze the performance of our relative pose classification method under varying amounts of visual overlap. 100\% overlap would indicate that two panoramas were captured in exactly the same position, with the scene unchanged between the two captures. On the other hand, 0\% overlap would indicate that the panoramas were captured in completely different locations, i.e. in two rooms, on opposite sides of a closed door (an example of an ``extreme'' baseline). We use a proxy metric, IoU of the texture map generated using HoHoNet-estimated \cite{Sun21cvpr_HoHoNet} monocular depth, which introduces some amount of noise.


\begin{figure}[t]
    \centering
    \includegraphics[trim=0 0 20 30, clip,width=0.22\columnwidth]{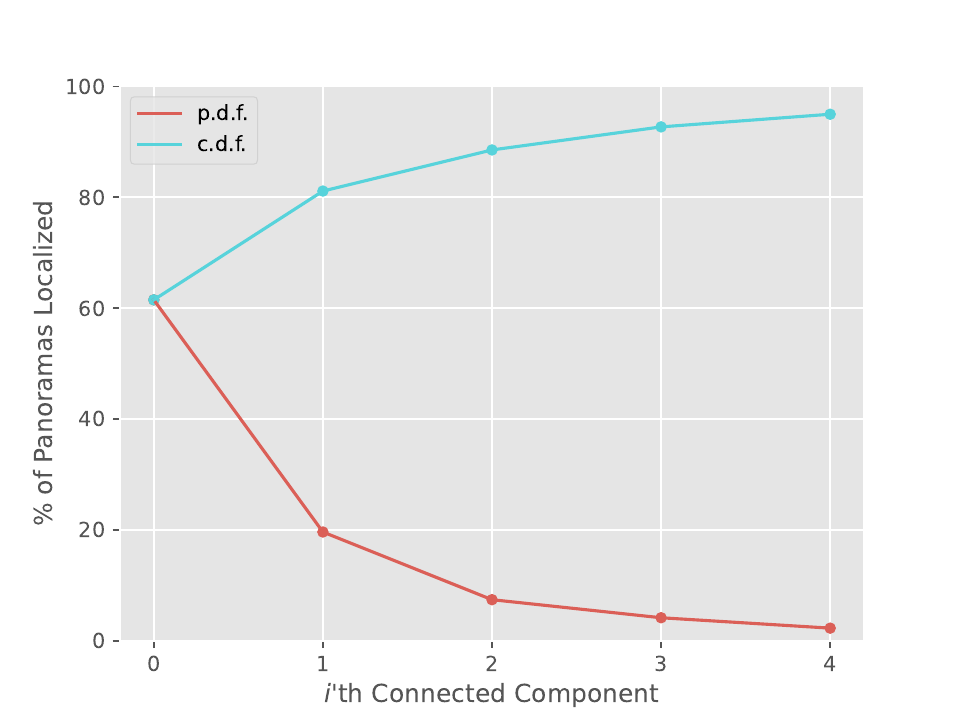} 
        \includegraphics[width=0.2\columnwidth]{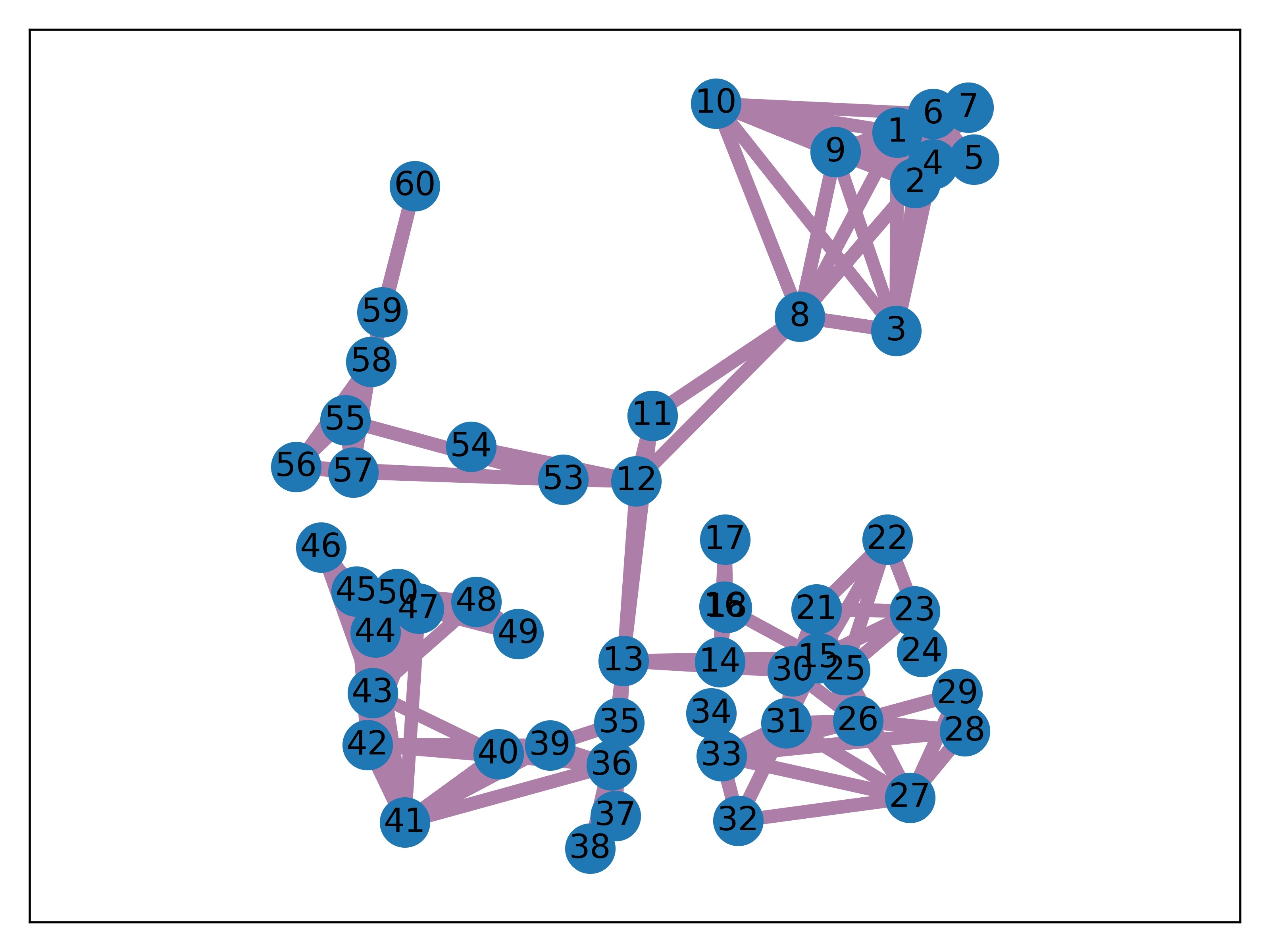}
        \includegraphics[width=0.2\columnwidth]{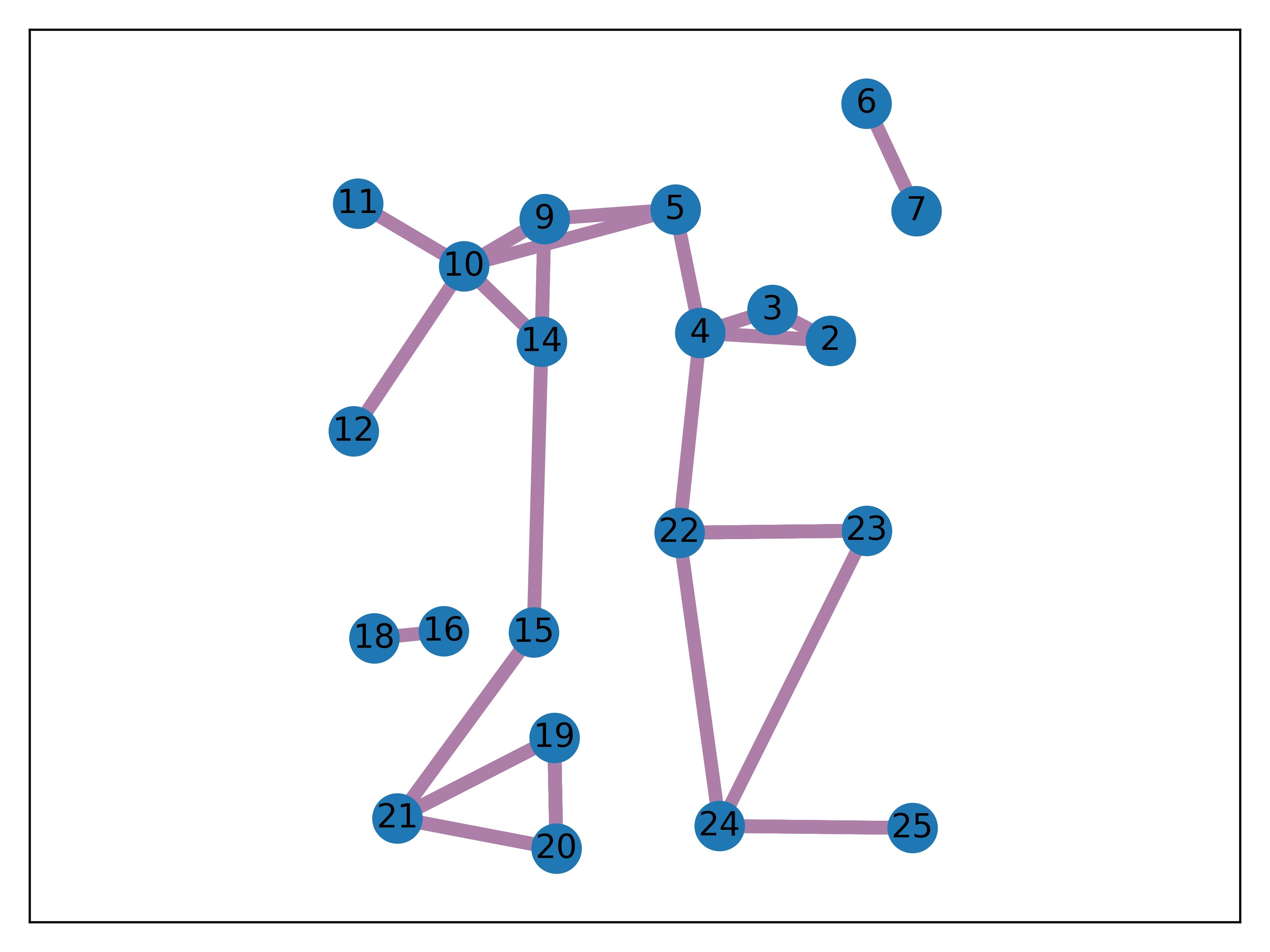}
        \includegraphics[width=0.2\columnwidth]{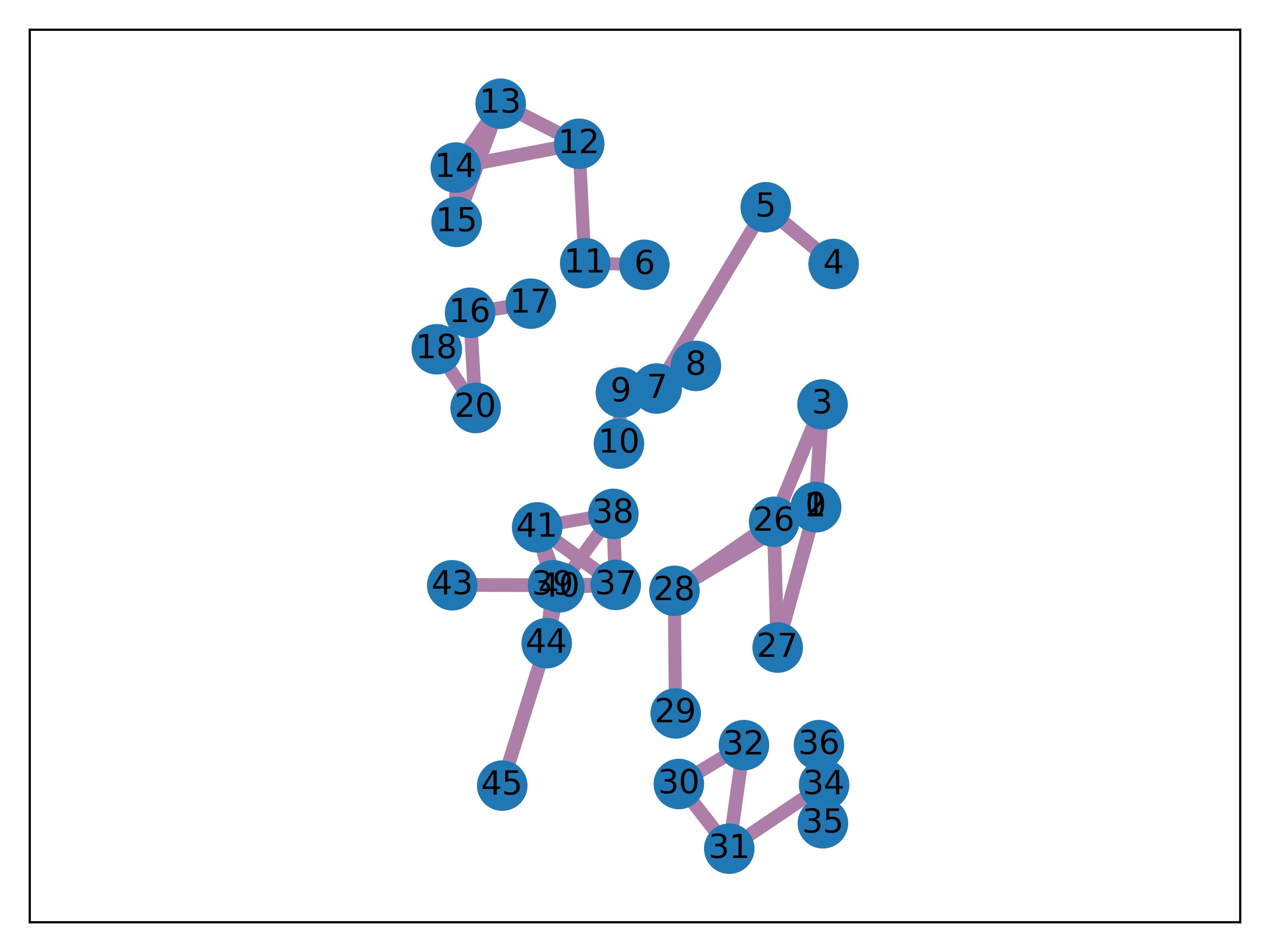}
    \caption{\emph{\textbf{Left:}} Distribution of localization percentage in the first 5 connected components, averaged over all test tours. \emph{\textbf{Right:}} Topology of global pose graphs for various different homes.}
    \label{fig:accuracy-vs-overlap-ablation}
    \vspace{-1.5mm}
\end{figure}


\subsection{Global Pose Estimation Results}
Next, we measure performance of both the ``front-end'' along with some form of global aggregation (``back-end''). We compare with two baselines from state-of-the-art structure from motion systems that support optimization from $360^\circ$ images.\\

\noindent \textbf{OpenMVG \cite{Moulon13iccv_GlobalFusionSfM,Moulon16iwrrpr_OpenMVG}.} We use the recommended setting for $360^\circ$ image input, with incremental SfM using an upright SIFT feature orientation, an upright 3-point Essential matrix solver with A-Contrario RANSAC, following the planar motion model described by \cite{Aly12wacv_StreetviewIndoors,Oskarsson18jmiv_TwoViewOrthographicEpipolarGeometry,Choi18ivc_MinimalRelativePosePlanarMotion}, with an angular constraint for matching. \\

\noindent \textbf{OpenSfM \cite{Gargallo16github_OpenSfM}.} Incremental SfM system that uses the Hessian-Affine interest point detector \cite{Mikolajczyk04ijcv_ScaleAffineInvariant}, SIFT feature descriptor \cite{Lowe04ijcv_SIFT}, and RANSAC \cite{Fischler81acm_RANSAC}. \\

\begin{figure*}
    \centering
    \vspace{-5mm}
    \includegraphics[trim=0 240 0 0, clip,width=\columnwidth]{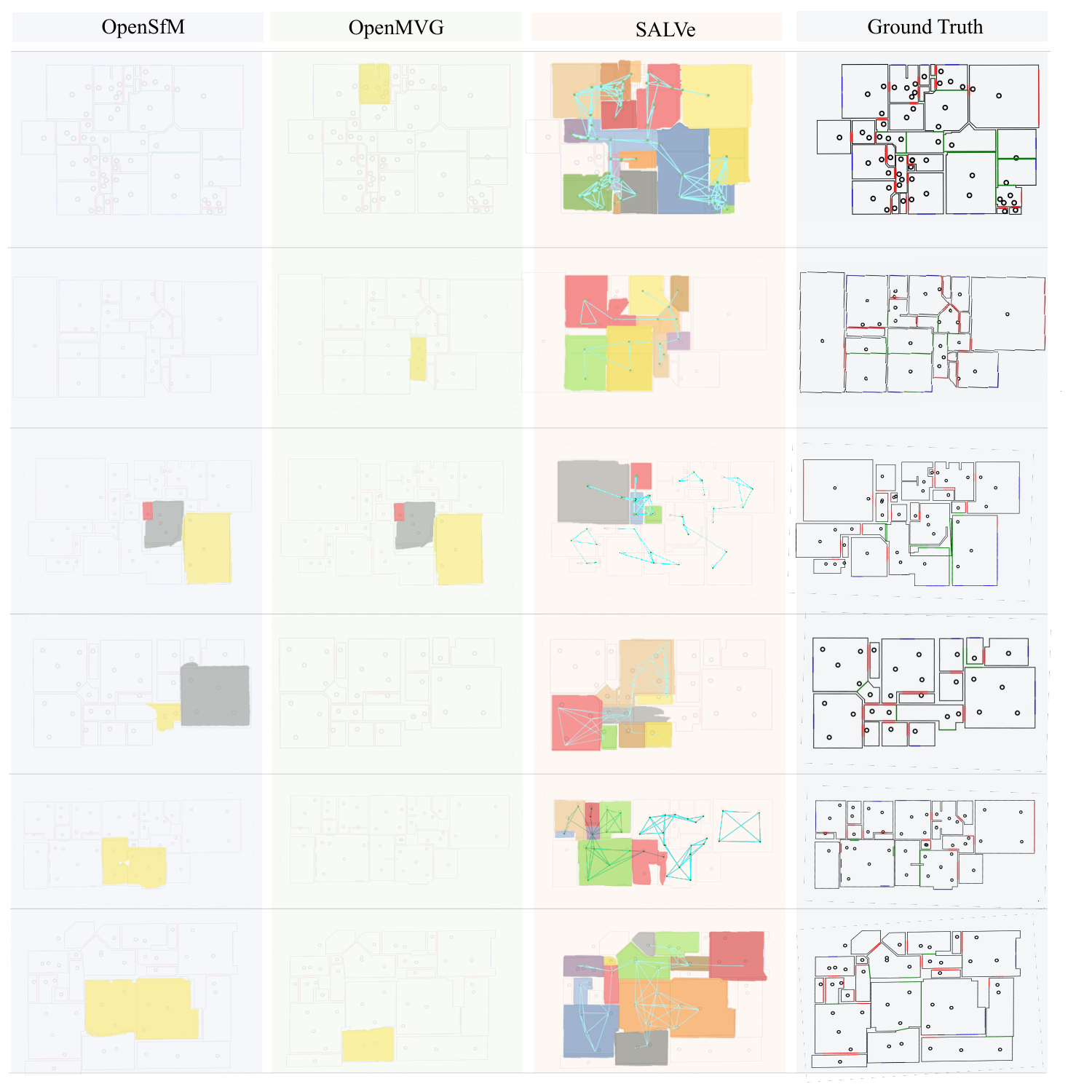}
    \vspace{-8mm}
    \caption{Qualitative comparison of floorplan results. \textbf{\emph{Column 1:}} OpenSfM. \textbf{\emph{Column 2:}} OpenMVG. \textbf{\emph{Column 3:}} Ours. \textbf{\emph{Column 4:}} Ground truth floorplan. All results are superimposed on the ground truth floorplan. Colored regions indicated the reconstruction result; at times, the baselines localize no panos. Our floorplan recall is significantly better than the state-of-the-art. Each row corresponds to a single floor of a different home. Colored lines represent W/D/O objects -- \textcolor{red}{doors},  \textcolor{darkpastelgreen}{openings} and \textcolor{blue}{windows}.  The multiple cyan edges in the overlaid graph correspond to verified W/D/O alignment hypotheses. For an open layout, a successful case often involves edges from panoramas in many different rooms to a single pano. These examples are intended to offer an even-handed selection of reconstructions that indicate both good performance as well as areas for improvements. Rows 1 and 6 illustrate good reconstructions.  Row 2 illustrates a more challenging case with only 1-2 panos in most rooms. Rows 3-5 are more challenging as they include bottlenecks in the actual physical layout, which is critical in joining connected components. } %
    \label{fig:qualresults}
\end{figure*}

In Table \ref{tab:zind-results-baselines}, we show the results of global pose estimation on the ZInD test set. We outperform OpenMVG by 656\% and OpenSfM by 257\% in the median percentage of panoramas localized (their 8.7\% and 22.2\% vs. our 57.1\%), with even lower median rotation error (our $0.17^{\circ}$  vs. their $0.37^{\circ}$ and $0.36^{\circ}$). Our median translation error is comparable (our 25 cm vs. their 12 cm and 10 cm). PGO is significantly more accurate than spanning tree when VP estimation is not employed (see Table \ref{tab:axis-alignment-ablation}). However, when using vanishing point-based dominant axis-alignment, both spanning trees and pose graph optimization on SALVe-verified measurements produce similar global aggregation results.  In the left column of Figure \ref{fig:qualresults}, we show the topological structure of the largest component of the pose graph for a few homes.

\begin{table}[t]
    \centering
    \caption{Results of global pose estimation on the ZinD test set. Two global aggregation methods are evaluated: spanning tree (`ST'), and pose graph optimization (`PGO'), with axis-alignment (`AA'). ST and PGO both use the same largest connected component of $\mathcal{G}$ as input, and thus localize an equal number of panoramas. }
    \vspace{-3mm}
    \begin{adjustbox}{max width=0.7\columnwidth}
	\begingroup
    \begin{tabular}{l|cc|cc|cc}
        \toprule 
        \textsc{\textbf{Method}} & \multicolumn{2}{|c|}{\textsc{\textbf{Localization \% }}} & \multicolumn{2}{|c|}{\textsc{\textbf{Tour Avg. Rotation  }}}       & \multicolumn{2}{|c}{\textsc{\textbf{Tour Avg. Translation }}} \\
                        & \multicolumn{2}{|c|}{}                                 & \multicolumn{2}{|c|}{\textsc{\textbf{Error (deg.)}}} & \multicolumn{2}{|c}{\textsc{\textbf{Error (meters)}}} \\
                        &  \textsc{mean} & \textsc{median} & \textsc{mean} & \textsc{median}             & \textsc{mean} & \textsc{median} \\
        \midrule
        \textsc{OpenSfM} \cite{Gargallo16github_OpenSfM} & 27.62 & 22.22 & 9.52 & 0.36 & 1.88 & 0.12 \\
        \textsc{OpenMVG} \cite{Moulon13iccv_GlobalFusionSfM,Moulon16iwrrpr_OpenMVG} & 13.94 & 8.70 & 3.84 & 0.37 & \textbf{0.41} & \textbf{0.10} \\
        \midrule
        \textsc{Ours (w/ ST + AA)} & \textbf{60.70} & \textbf{57.10} & \textbf{3.69} & \textbf{0.03} & 0.81 & 0.26  \\ 
        \textsc{Ours (w/ PGO + AA)} & \textbf{60.70} & \textbf{57.10} & 3.73 & 0.17 & 0.80 &	0.25    \\ 
        \bottomrule
    \end{tabular}
    \endgroup
    \end{adjustbox}
    \label{tab:zind-results-baselines}
    \vspace{-5mm}
\end{table}

\section{Discussion}

\noindent \textbf{Is deep learning necessary for verification, or can heuristics be used?} 
To verify pairwise alignment, matching texture is necessary but hard to feature engineer. Using geometry alone is insufficient (See Figure \ref{fig:binary-classification-pr-curve}(a-b) and Table \ref{tab:global-pose-estimation-ablation}), motivating others to explore graph neural networks for the task 
\cite{Shabani21iccv_ExtremeSfM}. We implemented rule-based baselines that classify BEV image pairs via FFT cross-correlation scores \cite{Reddy96tip_FftRegistration}, and found they do not work well due in part to difficulty in choosing thresholds. 
Previous works such as LayoutLoc \cite{Cruz21cvpr_ZillowIndoorDataset} have explored rule-based checking, but found that it only can be successful when given access to \emph{oracle} within-room pano grouping information; estimation of such within-room grouping (i.e. adjacency) is itself one of the fundamental challenges of global pose estimation in an indoor environment. 

\noindent \textbf{What type of semantic object is most useful for alignment in this semantic SfM problem?} Doors, but all are essential. Openings are the second-most effective object type to achieve complete localization, and windows are least effective. Among the alignments that the model predicts to be positives with confidence $\geq 97\%$, we find that 63\% originate from door-door hypotheses, 24\% originate from opening-opening hypotheses, and 20\% originate from window-window hypotheses. While rooms in residential homes are rarely connected by a window, these window alignments can provide additional redundancy, or ground alignments in very large open spaces when doors are not visible as in Fig.~\ref{fig:orthographic-samples}, pair 2. In Table \ref{tab:global-pose-estimation-ablation}, we report global pose estimation results when only one type of semantic object is used to create the edges $\mathcal{E}$ of the relative pose graph $\mathcal{G}$.

\begin{table*}[]
    \centering
    \caption{Results of ablation experiments on how inputs to SALVe affect global pose estimation accuracy and completeness. Pose graph optimization and vanishing point-based axis alignment (`PGO + AA') are utilized for all entries below. }
    \vspace{-3mm}
    \begin{adjustbox}{max width=\columnwidth}
	\begingroup
    \begin{tabular}{ccc | ccc     |cc|cc|cc}
        \toprule 
         \multicolumn{3}{c}{\textsc{\textbf{W/D/O Inputs}}}  & \multicolumn{3}{c}{\textbf{\textsc{Raster Inputs}}} & \multicolumn{2}{|c|}{\textsc{\textbf{Localization \%}}} & \multicolumn{2}{|c|}{\textsc{\textbf{Tour Avg. Rotation  }}}       & \multicolumn{2}{|c}{\textsc{\textbf{Tour Avg. Translation }}} \\
               Doors & Windows & Openings & Floor & Ceiling & Layout & \multicolumn{2}{|c|}{}                                 & \multicolumn{2}{|c|}{\textsc{\textbf{Error (deg.)}}} & \multicolumn{2}{|c}{\textsc{\textbf{Error (meters)}}}  \\
              & &  & Texture &   Texture &        &  \textsc{mean} & \textsc{median} & \textsc{mean} & \textsc{median}             & \textsc{mean} & \textsc{median}  \\
              \midrule
     \checkmark & \checkmark & \checkmark & \checkmark & \checkmark & & 60.70 & 57.14 &	3.73 & 0.17 & 0.80 & 0.25 \\ 
     \checkmark &            &            & \checkmark & \checkmark & & 43.30 & 40.00 &	2.41 & 0.07 & 0.59 & 0.20 \\ 
                & \checkmark &            & \checkmark & \checkmark & & 15.57 & 13.33 &	2.20 & \textbf{0.00} & 0.74 & \textbf{0.11} \\ 
                &            & \checkmark & \checkmark & \checkmark & & 23.87 & 23.08 	& \textbf{0.66} & 0.05 & \textbf{0.34} & 0.18 \\ 
    \hline
     \checkmark & \checkmark & \checkmark & \checkmark &            & & 60.64 & 58.33 & 3.75 & 0.15 & 0.91 & 0.25 \\ 
     \checkmark & \checkmark & \checkmark &            & \checkmark & & \textbf{60.93} & \textbf{64.58} & 10.94 & 0.28 & 2.12 & 0.35  \\ 
     \checkmark & \checkmark & \checkmark &        &    & \checkmark & 19.19 & 16.67 & 3.43 & 0.03 & 0.53 & \textbf{0.11}  \\ 
    \bottomrule
    \end{tabular}
    \endgroup
    \end{adjustbox}
    \label{tab:global-pose-estimation-ablation}
    \vspace{-5mm}
\end{table*}


\noindent \textbf{To what extent is the pose graph shattered into multiple clusters?} Typically, the first three connected components contain 61\%, 20\%, and 7\% of all panoramas (See Figure \ref{fig:accuracy-vs-overlap-ablation}a). We measure the distribution of connected components (CCs), as global pose estimation relies upon a single CC (we use the largest), and we find that often the second and third largest CCs are also large, indicating the potential for merging, e.g. combining ideas from ~\cite{Shabani21iccv_ExtremeSfM} or \cite{Son16cvpr_GrowingConsensus}. We compute an average probability density function and cumulative density function by averaging per-floor distributions across the test set. \\

\noindent \textbf{Is the RGB photometric signal from panoramas actually necessary, as opposed to solely using geometric context?} Yes, the RGB texture is essential. In Table  \ref{tab:global-pose-estimation-ablation}, we show that using a layout-only rasterization as input to the CNN, instead of a photometric texture map, leads to severe performance degradation. \\

\noindent \textbf{Does floor or ceiling texture provide a more useful signal for alignment classification?} Floor texture. However, using both signals jointly improves performance. In Table  \ref{tab:global-pose-estimation-ablation}, we show the results of using as input to the network only the floor texture maps, or only the ceiling texture maps, as opposed to reasoning about both jointly.\\

\begin{table}[b]
    \centering
    \vspace{-4mm}
    \caption{Comparison of results with and without axis-alignment (`AA') of relative poses (via vanishing angles) before global aggregation.  The amount of panoramas localized is unaffected, as adjacency is maintained during the correction. For this comparison, `oracle' layouts are used to isolate the effect of pose error. With vanishing point (VP) information, the difference between PGO and Spanning Tree is not statistically significant (1 cm and $0.04^\circ$ error on average).}
    \vspace{-1mm}
    \begin{adjustbox}{max width=0.6\columnwidth}
	\begingroup
    \begin{tabular}{l|cc|cc|cc}
        \toprule 
        \textsc{\textbf{Method}} & \multicolumn{2}{|c|}{\textsc{\textbf{Tour Avg. Rotation  }}}       & \multicolumn{2}{|c}{\textsc{\textbf{Tour Avg. Translation}}} &  \multicolumn{2}{|c}{\textsc{\textbf{Floorplan }}}  \\
                        &  \multicolumn{2}{|c|}{\textsc{\textbf{Error (deg.)}}} & \multicolumn{2}{|c|}{\textsc{\textbf{Error (meters)}}} & \multicolumn{2}{|c}{\textbf{IoU}}     \\
                        &  \textsc{mean} & \textsc{median} & \textsc{mean} & \textsc{median}             & \textsc{mean} & \textsc{median} \\
        \midrule
        Spanning Tree      & 5.41          & 1.92 & 0.86 & 0.33 & 0.55 & 0.52 \\
        Spanning Tree + AA & \textbf{3.69} & \textbf{0.03} & 0.81 & 0.26 & \textbf{0.56} & 0.52  \\
        PGO                & 4.93          & 1.53 & 0.81 & 0.29 & \textbf{0.56} & 0.52  \\ 
        PGO + AA           & 3.73 & 0.17 & \textbf{0.80} & \textbf{0.25} & \textbf{0.56} & \textbf{0.53}  \\
        \bottomrule
    \end{tabular}
    \endgroup
    \end{adjustbox}
    \label{tab:axis-alignment-ablation}
    \vspace{-6mm}
\end{table}

\noindent \textbf{Is a Manhattan world assumption helpful?} For pose estimation, yes, but for shape estimation, no. Many rooms at critical junctures in the floorplan are non-Manhattan in shape, and `Manhattanizing' them would be destructive when chaining together. However, room organization in a home is usually tied to three dominant, orthogonal directions. In Table \ref{tab:axis-alignment-ablation}, we show that using vanishing point estimation to align relative poses up to a $15^\circ$ correction significantly improves both global pose estimation accuracy and slightly improves floorplan reconstruction accuracy. Both vanishing point relative rotation angle correction and pose graph optimization are effective means of decreasing the rotation error. In the Appendix we show how using ground truth layout (near-perfect shape) and W/D/O locations affects performance, as an upper-bound on performance of the first module in our system.



\begin{table}[t]
    \centering
    \caption{Floorplan reconstruction results against the ground truth manually annotated floorplan. Floorplan 2D IoU is measured in the bird's eye view. The IoU is measured on the largest connected component. `AA' represents axis-alignment. }
    \vspace{-2mm}
    \begin{adjustbox}{max width=0.7\columnwidth}
	\begingroup
    \begin{tabular}{l| cc | cc| cc}
    \toprule
        \textsc{\textbf{Method}} & \multicolumn{2}{c}{\textsc{\textbf{Global Poses}}} & \multicolumn{2}{c}{\textsc{\textbf{Layout}}} & \multicolumn{2}{c}{\textsc{\textbf{Floorplan IoU}}} \\
         & \textsc{Oracle} & \textsc{Estimated} & \textsc{Oracle} & \textsc{Estimated} & \textsc{mean} & \textsc{median} \\
         \midrule
        \textsc{OpenSfM} & & \checkmark & \checkmark &  & 0.29 & 0.26 \\
        \textsc{OpenMVG} & & \checkmark & \checkmark &  & 0.16 & 0.07 \\
        \textsc{Ours}    & \checkmark & & & \checkmark & \textbf{0.94} & \textbf{0.95} \\
        \textsc{Ours (PGO + AA)}    &  & \checkmark & \checkmark &  & 0.56 & 0.53 \\
        \textsc{Ours (PGO + AA)}    &  & \checkmark &  & \checkmark &  0.49 & 0.45 \\
    \bottomrule
    \end{tabular}
    \endgroup
    \end{adjustbox}
    \label{tab:floorplan-reconstruction}
    \vspace{-2mm}
\end{table}


\subsection{Floorplan Reconstruction Results}
\label{sec:floorplan-reconstruction-results}
Next, we compare performance of the entire floorplan reconstruction system. In Table \ref{tab:floorplan-reconstruction}, we demonstrate that compared to traditional SfM with oracle room layout and oracle scale, our end-to-end system is able to produce more accurate floorplans with estimated room layouts (our 0.49 mean IoU vs. OpenSfM's 0.29 and OpenMVG's 0.16). The 0.56 mean IoU score using our estimated global poses and oracle layout primarily reflects the completeness of our final floorplan. With oracle pose and estimated room layouts, the 0.94 mean IoU reflects the accuracy of our layout estimation and stitching stages. This baseline has significantly larger IoU in part because the `oracle' poses are provided for \emph{all} panoramas (see the Appendix for comparison visualizations).

\noindent \textbf{Qualitative Results.} Fig. \ref{fig:qualresults} provides qualitative results for a number of different homes. For floors of some homes, our method produces nearly complete reconstructions, while for others, the results are more sparse. As shown by the third column of Fig. \ref{fig:qualresults}, the topology of the pose graph directly affects the completeness of the reconstruction; when multiple large connected components appear, the reconstruction is shattered apart. For several homes, OpenMVG and OpenSfM fail to converge, localizing no panoramas.



\section{Conclusion}

We present a new system for automatic 2D floorplan reconstruction from sparse, unordered panoramas. This work represents a breakthrough in the completeness of reconstructed floorplans, with over two times more coverage than previous systems \cite{Gargallo16github_OpenSfM,Moulon16iwrrpr_OpenMVG}, without sacrificing accuracy. 
We demonstrate how {\em SALVe}, our novel pairwise learned alignment verifier, capitalizes on the mature field of semantic detection of features (W/D/O) to handle a tractable number of alignment hypotheses and generate high-quality results. A human annotator may use it to accelerate labeling by automatically generating the majority of necessary decisions before making the final choices about glueing connected components. Fig. \ref{fig:qualresults} only illustrates the largest CC; other CCs are also generated, but not shown (Fig. \ref{fig:accuracy-vs-overlap-ablation}, a CDF of 89\% for the first 3 CCs).




\noindent \textbf{Limitations.} Because the number of pairwise alignments is combinatorial in the number of W/D/O, the runtime of the current system is limited, although we have not heavily optimized it. As ZInD \cite{Cruz21cvpr_ZillowIndoorDataset} contains only unfurnished homes, our system has not yet been evaluated in a furnished home regime, due to dataset availability. Camera localization completeness is still in the 55-60\% range. With future improvements to each part of the system, especially omnidirectional depth estimation and layout estimation, we expect floorplan reconstruction performance to continue to improve.


\clearpage
%
%
\bibliographystyle{splncs04.bst}
\bibliography{egbib}

\appendix
\section{Appendix}

In this Appendix, we provide additional analysis and implementation details. In Section A.1, we provide qualitative comparisons of our floorplan reconstructions, vs. an upper-bound \emph{oracle} baseline that uses ground-truth global pose estimation. In Section A.2, we provide quantitative analysis of SALVe's relative pose classification accuracy with various input modalities. In Section A.3, we provide pseudo-code for our layout stitching algorithm. In Sections A.4 and A.5, we report detailed quantitative analysis of W/D/O detection accuracy, and W/D/O and layout estimation failure cases. In Section A.6, we describe the coordinate systems used in our work. In Sections A.7, A.8, and A.9, we provide additional implementation details about rendering, vanishing-point based axis alignment, and pose graph optimization and spanning tree aggregation. In Section A.10, we describe ablation experiments that compare the use of ground truth W/D/O and ground truth layout, vs. estimated W/D/O and estimated layout. In Section A.11, we provide additional discussion about our evaluation procedures versus those of concurrent work \cite{Shabani21iccv_ExtremeSfM}. In Sections A.12-15, we provide additional analysis and further examples of positive and negative training examples. In Section A.16, we discuss additional details about verifier training and data augmentation, and in Section A.17, we discuss ethical concerns associated with the work

\subsection{Qualitative Results: Predicted vs. Oracle Poses}
\label{sec:oracle-poses-qual-results}
\begin{figure}
    \centering
    \includegraphics[trim=20 10 10 36, clip,width=\columnwidth]{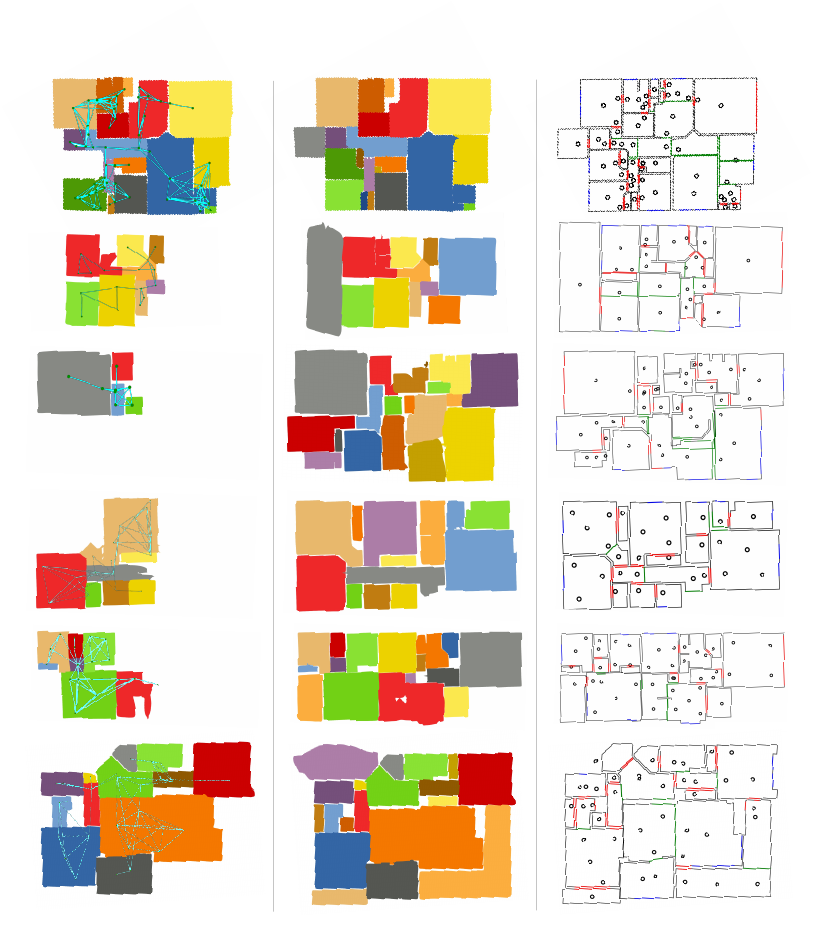}
    \caption{Example floorplan results of varying completeness, comparing SALVe’s performance vs. an upper bound (perfect global pose estimation). \emph{\textbf{Left:}} predicted poses of the largest connected component and predicted room layout. \emph{\textbf{Middle:}} oracle (ground truth) poses and predicted room layout. \emph{\textbf{Right:}} ground truth floorplan with positions of captured panoramas. }
    \label{fig:reconstruct-pred-vs-oracle-poses1}
\end{figure}

\begin{figure*}
    \centering
    
    \includegraphics[width=0.9\columnwidth]{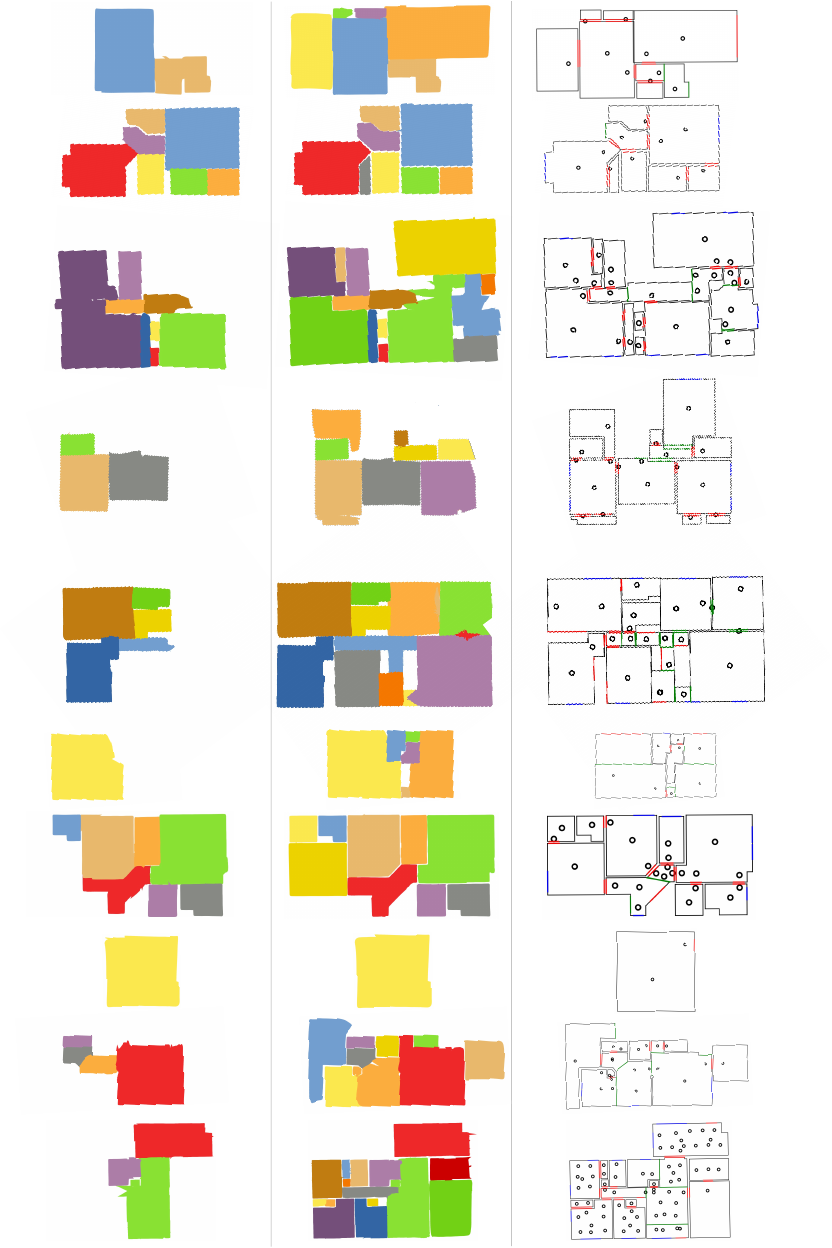}
    \caption{Additional comparison between SALVe's performance and an upper bound (perfect global pose estimation). Each row corresponds to a single floor of a different home. \emph{\textbf{Left:}} predicted poses of the largest connected component and predicted room layout. \emph{\textbf{Middle:}} oracle poses and predicted room layout. \emph{\textbf{Right:}} ground truth floorplan with positions of captured panoramas. Colored lines represent W/D/O objects -- \textcolor{red}{doors},  \textcolor{darkpastelgreen}{openings} and \textcolor{blue}{windows}.}
    \label{fig:reconstruct-pred-vs-oracle-poses2}
\end{figure*}

In this section, we provide qualitative comparisons with a baseline that stitches predicted layouts placed at `oracle' global pose locations (referred to by Section \ref{sec:floorplan-reconstruction-results} of the main paper). For this baseline, the high fidelity of reconstructed shapes (middle column of Fig. \ref{fig:reconstruct-pred-vs-oracle-poses1} and Fig. \ref{fig:reconstruct-pred-vs-oracle-poses2}) demonstrates the maturity of modern layout estimation networks. This baseline also illustrates the impact of global pose estimation on the entire system.

\subsection{Additional Analysis of Relative Pose Classification Accuracy}
\label{sec:rel-pose-classification-acc-input-modalities}
Here we provide a more comprehensive quantitative analysis of the influence of input modalities and CNN backbone architecture on SALVe's relative pose classification accuracy (referred to in Section \ref{sec:relative-pose-classification-results-main-paper} of the main paper). We compare ceiling-only texture map input, vs. floor-only texture map input, vs. using both as input.

\begin{table}
    \centering
    \caption{Relative pose classification accuracy on the ZInD test split with different inputs and architectures. Precision, recall, and mean accuracy are reported. Extreme class imbalance means that with more expressive model architectures, gains in mean accuracy are minor, but gains in precision are significant.}
    \begin{adjustbox}{max width=0.5\columnwidth}
	\begingroup
    \begin{tabular}{c c c c c c}
    \toprule
        \textsc{Model } & \textsc{Ceiling } & \textsc{Floor } & \textsc{Prec.} & \textsc{Rec.} & \textsc{mAcc.} \\
        \textsc{ Architecture} & \textsc{ Texture Map} & \textsc{Texture Map} &  &  &  \\
        \midrule
        \textsc{ResNet-50} & \checkmark & \checkmark  & 0.77 & 0.91 & 0.96 \\
        \textsc{ResNet-152} & \checkmark & \checkmark  & \textbf{0.85} & \textbf{0.91} & \textbf{0.95} \\
        \textsc{ResNet-152} & \checkmark &             & 0.70 & 0.88 & 0.93 \\
        \textsc{ResNet-152} &            & \checkmark  & 0.84  & \textbf{0.91}  & \textbf{0.95} \\
        \bottomrule
    \end{tabular}
    \endgroup
    \end{adjustbox}
    \label{tab:rel-pose-classification}
    \vspace{-4mm}
\end{table}

\subsection{Details on Layout Stitching for Floorplan Reconstruction}
\label{sec:layout-stitching-algo-pseudo-code}
This section provides additional details about the reconstruction algorithm mentioned in Section \ref{sec:floorplan-reconstr-approach} and Figure \ref{fig:floor-plan-stitching} of the main paper.

\begin{figure}
    \vspace{-10mm}
    \centering
    \subfloat[]{
        \includegraphics[trim=0 210 160 0,clip,width=0.25\columnwidth]{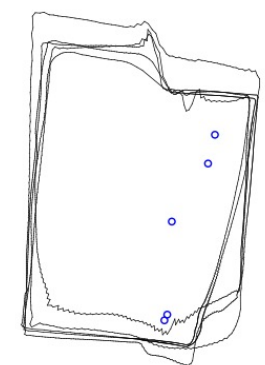} 
    }
    \subfloat[]{
        \includegraphics[trim=150 210 20 0,clip,width=0.25\columnwidth]{latex/figs/shape_post_processing/shape_post_processing_v2.pdf} 
    }
    \subfloat[]{
        \includegraphics[trim=0 20 160 200,clip,width=0.25\columnwidth]{latex/figs/shape_post_processing/shape_post_processing_v2.pdf} 
    }
    \subfloat[]{
        \includegraphics[trim=140 20 20 200,clip,width=0.25\columnwidth]{latex/figs/shape_post_processing/shape_post_processing_v2.pdf} 
    }
    \caption{Visualization of room shape reconstruction using localized panoramas grouped by room. \textbf{(a)} Predicted room layout and predicted panorama locations \textcolor{blue}{(blue dots)}. \textbf{(b)} Predicted room layout with contour confidence (transparency) and predicted panorama locations (blue dots). \textbf{(c)} Overlay of room contours generated by voting on the highest confidence contour point at each panorama column from each panorama view. The final room layout is the union of these view-dependent contours of highest confidence. \textbf{(d)} Ground truth room shape and ground truth panorama positions.}
    \label{fig:room-layout-processing}
\end{figure}

Floorplan reconstruction involves three steps: (1) panorama room grouping, (2) highest confidence room contour extraction, and (3) floorplan stitching. Please see Algorithm \ref{alg:floorplan-reconstruction} for implementation details. In Figure \ref{fig:room-layout-processing}, we demonstrate the process of generating final room layout using estimated panorama poses grouped by step (1). Comparing Fig \ref{fig:room-layout-processing}(b) to Fig. \ref{fig:room-layout-processing}(a), we can see that room contour confidence provide useful guidance in selecting the high confidence contour point among different views. In Fig. \ref{fig:room-layout-processing}(c), each view-dependent room contour largely will agree with each other. In the end, we take the union of different view-dependent room contours to account for the occlusions from each panorama view.

\begin{algorithm}
\caption{Floorplan Reconstruction for a Connected Component in Pose Graph}
\label{alg:floorplan-reconstruction}
\begin{algorithmic}
\State \textbf{Inputs:}
$\{I_{i}\}$: A list of the input panorama images in the connected component. \\ 
$\{\mathbf{T}_{i}\}$: Estimated panorama poses from pose graph, in top-down global 2D coordinates. \\ 
$\{(C_{i}, \sigma_{i})\}$: Estimated room contour points and contour point confidence for panorama $I_{i}$, in top-down global 2D coordinates.  (One point per panorama column.) \\

\State \textbf{Output:} 
\State $S_{floorplan}^{opt}$: Optimized floor plan polygon shape. \\

\State \textbf{Solution:} 
\State \textcolor{gray}{\% Step 1: Group panoramas that come from the same room.}
\State Initialize panorama connectivity graph $\Gamma$ with one node per pano $I_{i}$ and no edges.
\For{$(\mathbf{T}_{i}, \mathbf{T}_{j}) \in \{\mathbf{T}_{i}\} \times \{\mathbf{T}_{i}\}$}
\State ${IoU} \gets ComputeContourIoU(\mathbf{T}_{i}, C_{i}, \mathbf{T}_{j}, C_{j})$
\If{${IoU} > Threshold$}
    \State $\Gamma.AddEdge(i, j)$
\EndIf
\EndFor
\State \textcolor{gray}{\% Each connected component in $\Gamma$ is a room.}
\State $\mathcal{G} = \{G^r, r = 1, ..., N_{rooms}\} \gets \Gamma.GetConnectedComponents()$ \\

\State \textcolor{gray}{\% Step 2: Extract highest confidence contour for each room.}
\For {$G^r \in \mathcal{G}$:}

\State Let optimized room shape $S_{r}^{opt} = \emptyset$
\For{$I_{i} \in G^r$}
\State $\mathcal{P}^i = \{(P^{i}_{j},\sigma^{i}_{j})\}\; \forall \; I_j \in G^r$, where $(P^{i}_{j}, \sigma^{i}_{j})$ are the projections of $(C_{j}, \sigma_{j})$ onto pano $i$'s image
\State In each image column of pano $i$, choose the most confident contour point from $\mathcal{P}^i$.
\State  $S^{i} \gets$ the selected points, projected back into the 2D global coordinates using $\mathbf{T}_{i}$



\EndFor
\State $S^{opt}_{r}=\bigcup_{I_i \in G^r} polygon(S^{i})$
\EndFor \\

\State \textcolor{gray}{\% Step 3: Floorplan stitching}
\State $S_{floorplan}^{opt}=\bigcup_{l=0}^{N_{rooms}} S_{r}^{opt}$ 

\end{algorithmic}
\end{algorithm}

\subsection{Details on W/D/O Detection Evaluation}
\label{sec:horizonnet-detection-acc-appendix}
In Section \ref{sec:layout-wdo-estimation-acc-main-paper} of the main paper, we report W/D/O detection results of our HorizonNet \cite{Sun19cvpr_HorizonNet} model at a 70\% IoU threshold. For completeness, we provide here an evaluation of 1d IoU at 50\%, 70\%, and 90\% true positive thresholds (see Table \ref{tab:wdo-detection-all-iou-thresholds}).



\begin{table}[]
    \centering
    \vspace{-5mm}
    \caption{Additional W/D/O detection accuracy results.}
    \begin{adjustbox}{max width=\columnwidth}
	\begingroup
    \begin{tabular}{l | lll | lll | lll}
\toprule
        & \multicolumn{3}{|c|}{0.5 IoU} & \multicolumn{3}{|c|}{0.7 IoU} & \multicolumn{3}{|c}{0.9 IoU} \\
        & Prec.    & Rec.    & F1     & Prec.    & Rec.    & F1     & Prec.    & Rec.    & F1     \\
        \midrule
\textsc{Door}    & 0.88     & \textbf{0.92}    & 0.90    & 0.87    & \textbf{0.91}    & 0.89   & 0.86    & 0.81    & 0.84   \\
\textsc{Window}  & \textbf{0.94}     & 0.91    & \textbf{0.92}   & \textbf{0.94}    & 0.89    & \textbf{0.91}   & \textbf{0.93}    & \textbf{0.82}    & \textbf{0.87}   \\
\textsc{Opening} & 0.79     & 0.65    & 0.72   & 0.78    & 0.59    & 0.67   & 0.72    & 0.43    & 0.54  \\
    \bottomrule
    \end{tabular}
    \endgroup
    \end{adjustbox}
\label{tab:wdo-detection-all-iou-thresholds}
\end{table}

\subsection{Layout and W/D/O Failure Cases}
\label{sec:horizonnet-failure-cases}
\begin{figure}[!h]
    \centering
    \subfloat[]{
        \includegraphics[width=0.25\columnwidth]{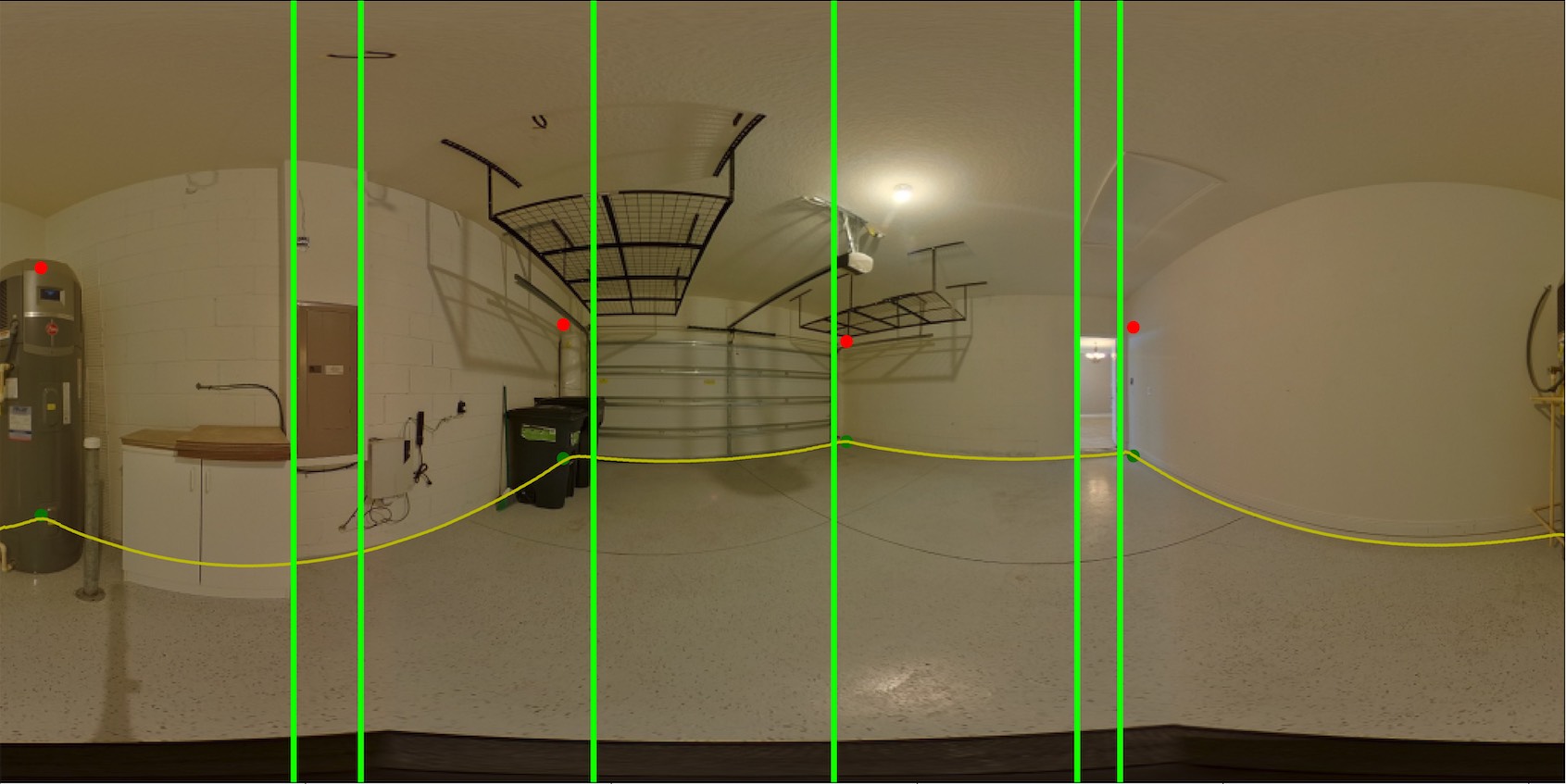}
        \includegraphics[width=0.25\columnwidth]{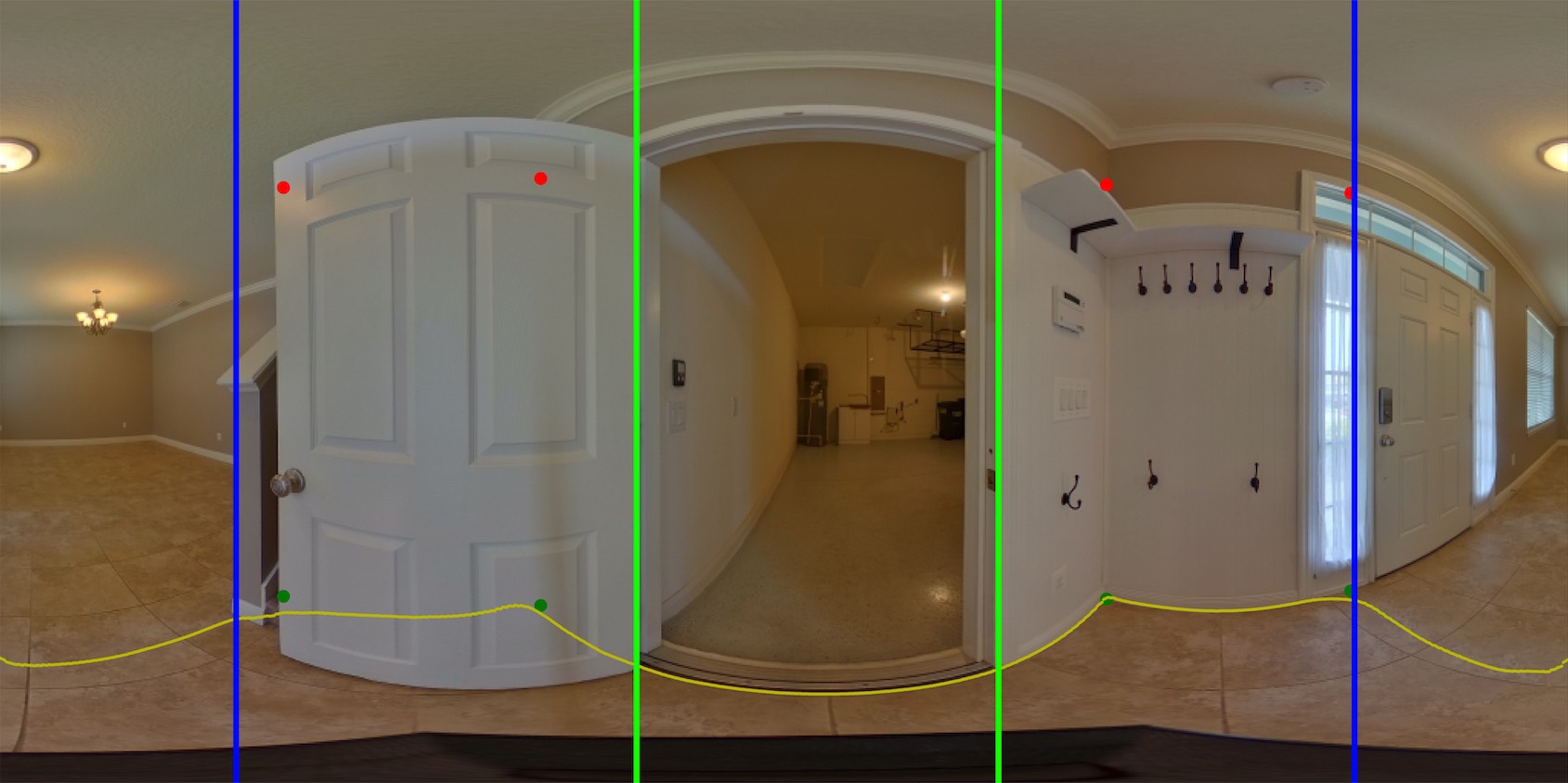}
    }
    \subfloat[]{
        \includegraphics[width=0.25\columnwidth]{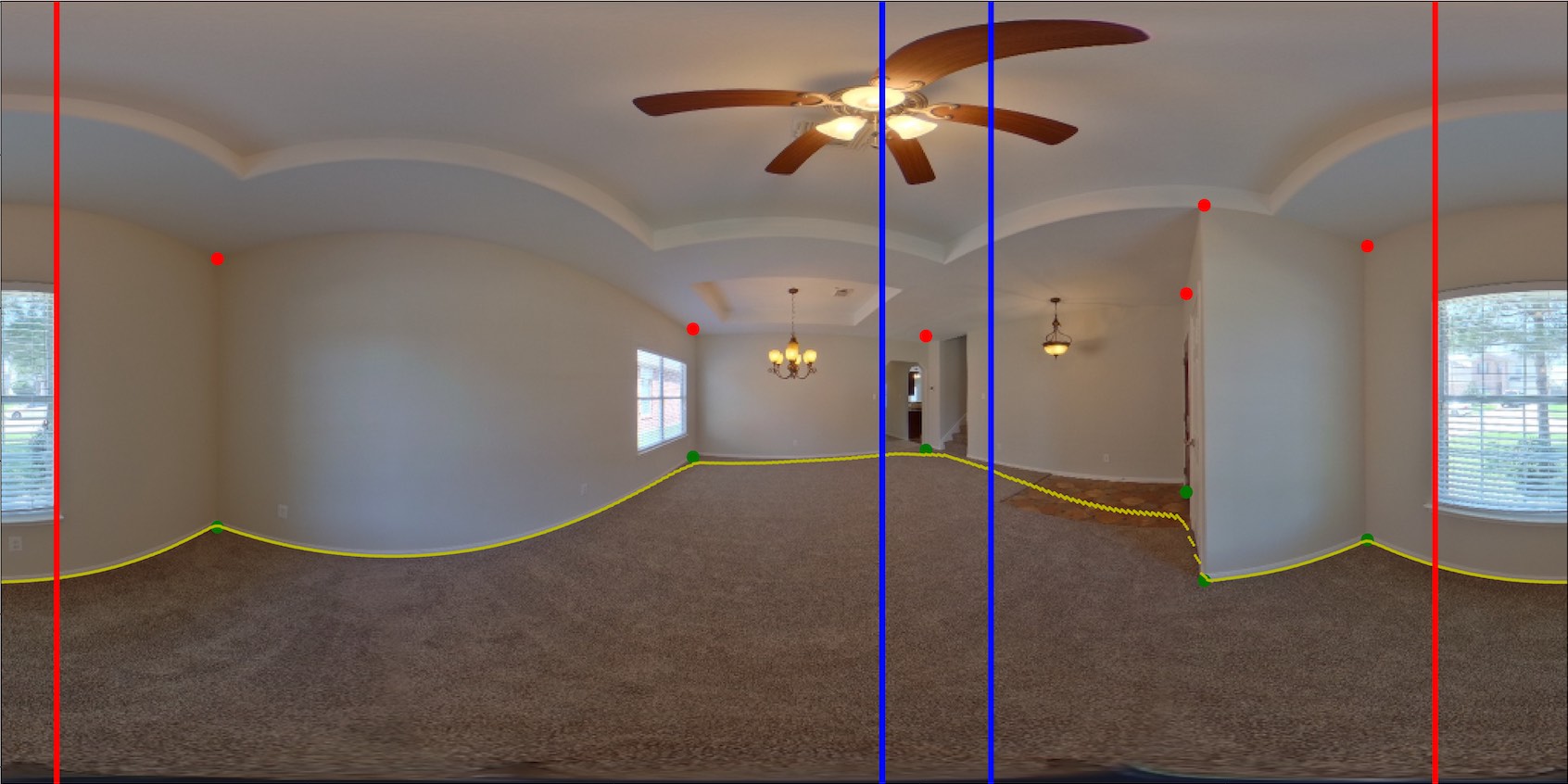}
        \includegraphics[width=0.25\columnwidth]{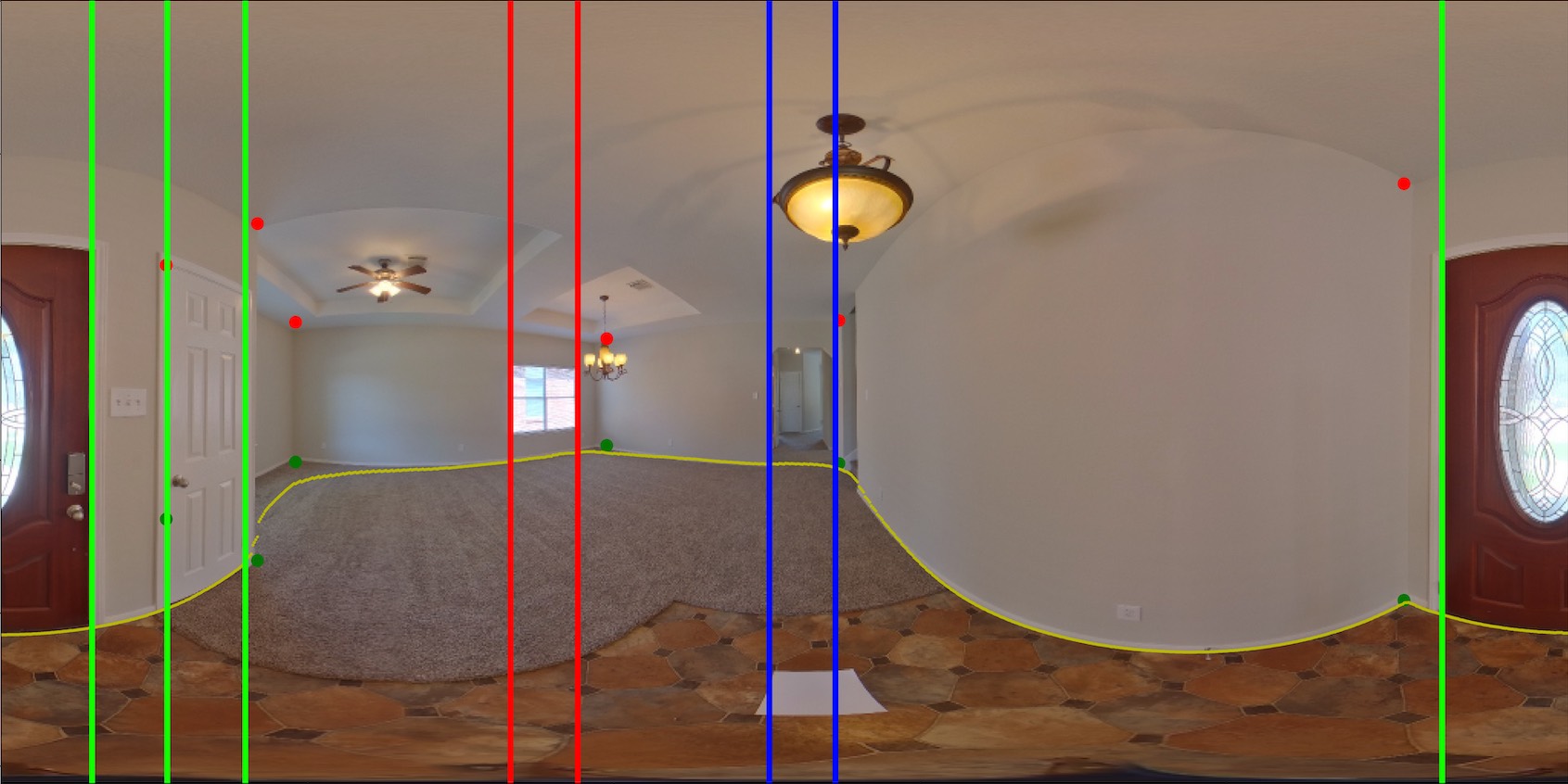}
    }
    \vspace{-2mm}
    \caption{Mistakes made by the joint HorizonNet + W/D/O model. Vertical lines indicate start and end columns for each W/D/O object --
    \textcolor{red}{window}, \textcolor{darkpastelgreen}{door}, and \textcolor{blue}{opening}. The yellow contour indicates the predicted floor-wall boundary, and dots indicate corner predictions (floor-wall corners in green, and ceiling-wall corners in red). Left and right images are panoramas across which we seek to match W/D/O objects. \emph{\textbf{Top:}} A circuit breaker panel is mistakenly identified as a door \textbf{(top left)}, but redundancy still allows matching of the true garage door. This allows a relative pose hypothesis to be generated between the foyer and garage panoramas, that have very little visual overlap. \emph{\textbf{Bottom:}} A false negative window prediction and inaccurate opening prediction \textbf{(bottom left)} makes matching with the \textbf{(bottom right)} panorama impossible using W/D/O.}
    \label{fig:horizonnet-mistakes}
\end{figure}

Section 5.4 of the main paper discusses the accuracy of W/D/O detection. Here we offer, in Figure \ref{fig:horizonnet-mistakes}, two examples of some the failure modes of the layout and W/D/O estimation model that provides the input to SALVe.  

\subsection{Coordinate System Conventions}
\label{sec:zind-coord-systems}
\begin{figure*}[!h]
    \centering
    \subfloat[Panoramic Spherical]{
        \includegraphics[trim=400 50 100 100,clip,width=0.3\columnwidth]{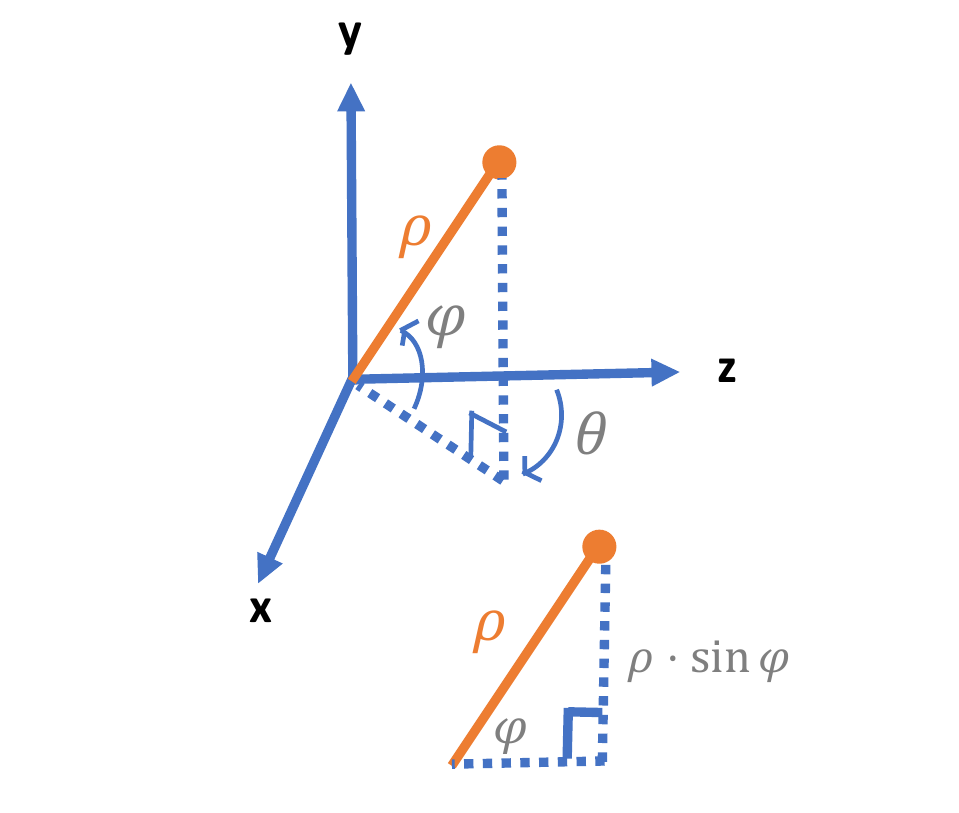}
    }
    \subfloat[Room Cartesian]{
        \includegraphics[trim=400 50 100 100,clip,width=0.3\columnwidth]{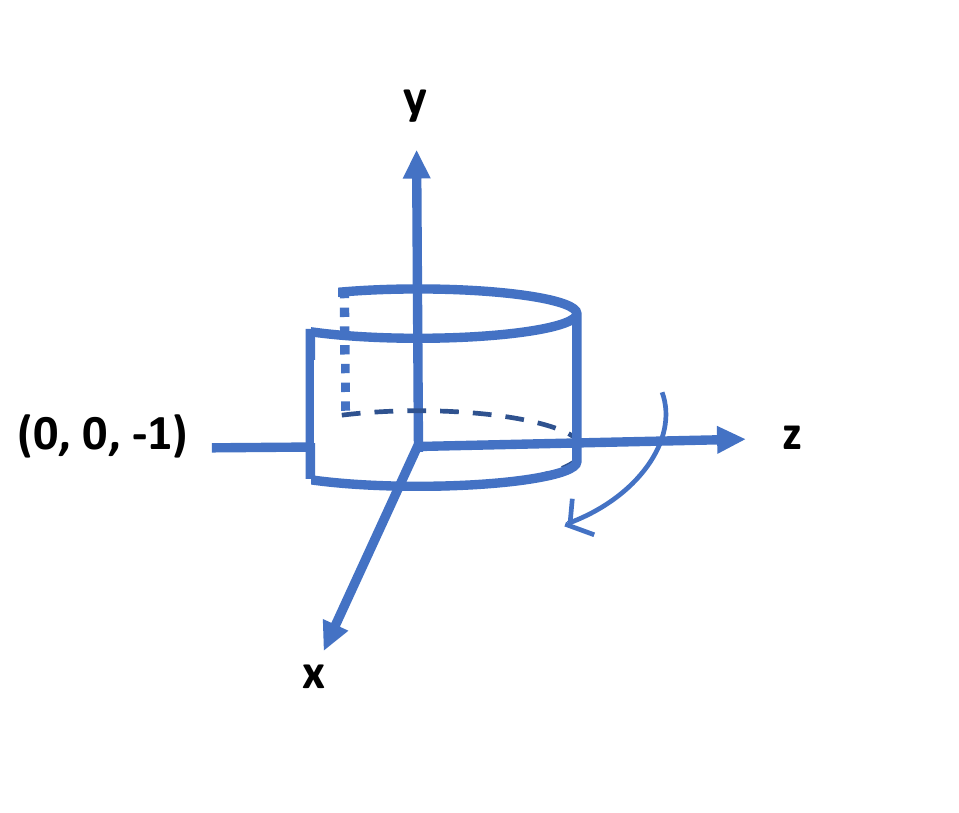}
    }
    \subfloat[World-metric Cartesian]{
        \includegraphics[trim=400 70 100 100,clip,width=0.3\columnwidth]{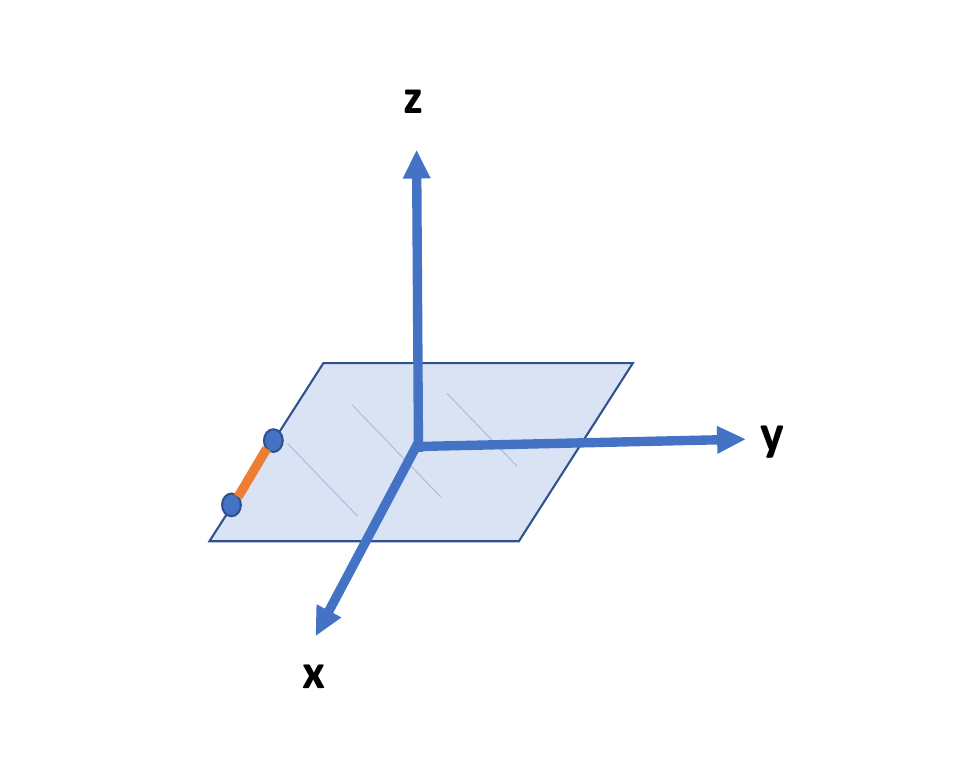}
    }
    \vspace{2mm}
    \subfloat[]{
        \includegraphics[trim=100 10 300 10,clip,width=0.27\columnwidth]{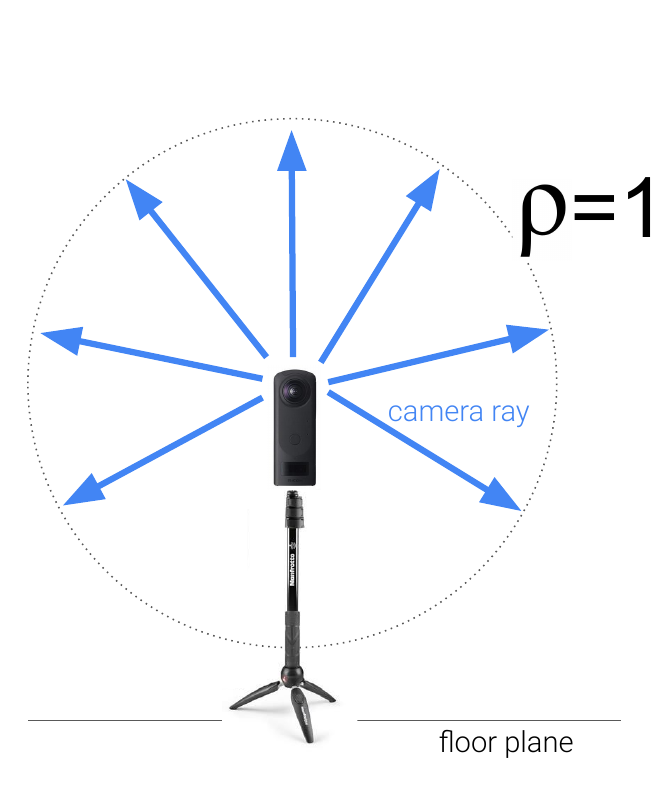}
    }
    \subfloat[]{
        \includegraphics[trim=50 100 50 90,clip,width=0.5\columnwidth]{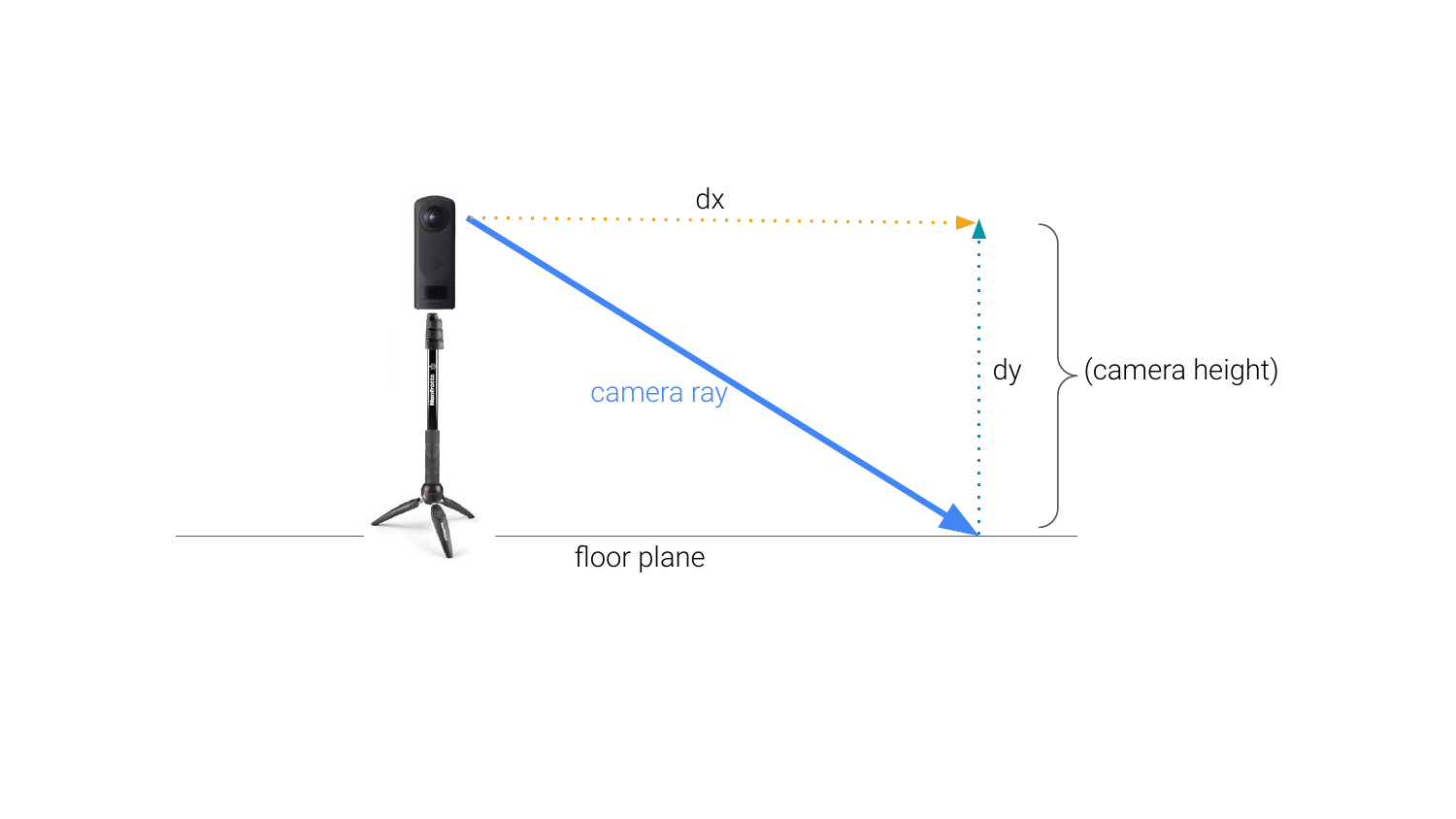}
    }
    \caption{Coordinate system conventions. For (b), note that an equirectangular projection of a panorama actually captures a sphere, not a cylinder.}
    \label{fig:coordsystems}
\end{figure*}

Figure~\ref{fig:coordsystems} shows the coordinate systems used in our work: panoramic spherical coordinate system, room Cartesian coordinate system, world-normalized Cartesian coordinate system (with camera height set to 1.0), and the world-metric coordinate system. These are also the coordinate conventions used by ZInD\footnote{ZInD is publicly available under the following license: \url{https://bridgedataoutput.com/zillowterms}. }.

\noindent \textbf{Scaling to Metric Space.} With known camera height, a predicted floor-wall boundary in pixel space with vertices $\{(u,v)_k \}_{k=1}^{K}$ can be mapped to 3D by first converting each vertex to spherical coordinates, and then to Cartesian coordinates, as follows:
\begin{equation}
\begin{aligned}
    \theta &= \Big(u \cdot \frac{2 \pi}{(w-1)}\Big) - \pi, \hspace{5mm} \theta \in [-\pi,\pi] \\
    \varphi &= \pi \Big(1 - \frac{v}{(h-1)}\Big) - \frac{\pi}{2},\hspace{5mm} \varphi \in \Big[-\frac{\pi}{2},  \frac{\pi}{2}\Big] ,
\end{aligned}
\end{equation}
where $h$ and $w$ is the height and width of input image in pixels.

We next obtain ray directions $(x,y,z)$ in Cartesian space by assuming that all points $(\theta,\phi,\rho)$ lie on the unit sphere (see Figure \ref{fig:coordsystems}d), i.e., $\rho=1$, and 
\begin{equation}
\begin{aligned}
   x &= \cos(\varphi) \sin(\theta) \\
   y &= \sin(\varphi) \\
   z &= \cos(\varphi) \cos(\theta) .
\end{aligned}
\end{equation}
Finally, we rescale the length of each ray such that it intersects the ground plane at $y=0$, i.e., the magnitude of its $y$ coordinate is equal to the camera height $h_c$ (see Figure \ref{fig:coordsystems}e). These rescaled ray directions are now coordinates in meters.

\subsection{Texture Mapping Procedure}
\label{sec:texture-mapping-procedure}
In this section, we discuss the procedure we use when creating bird's eye view (BEV) texture maps, as mentioned in Section \ref{sec:impl-details-main-paper} of the main paper.

When texture mapping an orthographic view using the monocular estimated depth map from ~\cite{Sun21cvpr_HoHoNet}, we use all 3D points $\geq 1$m below camera for rendering the floor, and all points $\geq 0.5$m above the camera for rendering the ceiling. We render a $10 \times 10$m region, using a resolution of 0.02 m/pixel, creating a $500 \times 500$ image.

\begin{figure}
    \centering
    \includegraphics[width=0.4\columnwidth]{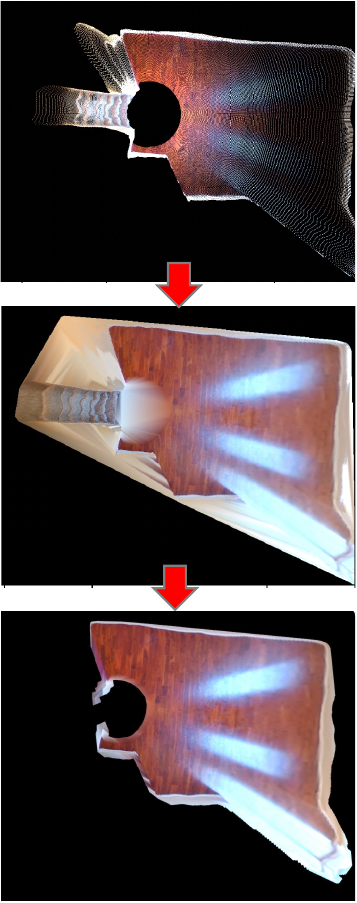}
    \caption{Visualization of the sparse to dense interpolation scheme. \textbf{\emph{Top}:} sparse texture map from mono-depth. \textbf{\emph{Middle}:} linearly interpolated texture map. \textbf{\emph{Bottom}:} result after removing interpolation artifacts.}
    \label{fig:interpolation}
\end{figure}

Afterwards, we apply dense interpolation to the initial texture map. When generating the orthographic imagery, the raw signal from pixel values at all backprojected depth map locations is sparse and insufficient. We rely upon interpolation to generate a dense canvas from the sparse canvas (see Figure \ref{fig:interpolation}, top). This interpolation also adds unwanted and undesirable interpolation artifacts (See Figure \ref{fig:interpolation}, middle subfigure). We design another step to identify the regions where the signal was too sparse to interpolate accurately. We convolve the canvas that is populated with sparse values with a box filter.  We zero-out portions of an interpolated image where the signal is unreliable due to no measurements. If a $K \times K$ subgrid of an image has no sparse signals within it, and is initialized to a default value of zero, then convolution of the subgrid with a box filter of all 1's will be zero. In short, if the convolved output is zero in any $ij$ cell, then we know that there was no true support for interpolation in this region, and we should mask out this interpolated value. We multiply the  interpolated image with binary unreliability mask to zero out unreliable values. Convolution with a large kernel, e.g., $11 \times 11$ pixels in size on a 500p image, can be done on the GPU. We populate the canvas from bottom to top.

\subsection{Vanishing Point Axis Alignment}
\label{sec:vp-axis-alignment}
Here we provide details about how we refine the hypothesized relative alignment described in Section \ref{sec:generate-alignment-hypotheses} of the main paper.

 To correct minor errors of W/D/O vertex localization, we compute vanishing points and convert them to a vanishing angle  $\theta_{vp}$ with direction voting from line segments \cite{Zhang14eccv_PanoContext}. The vanishing angle $\theta_{vp}$ is defined as the horizontal angle between left edge of panorama and the first vanishing point from the left side of the panoramic image. We then refine the panorama horizontal rotation by aligning the pair of vanishing angles, while maintaining the distance between the matching W/D/O. The angular adjustment can be represented by: $\theta_{correction} = (\theta_{vp,1} - \theta_{vp,2}) - {}^{2}\theta_{1}$, where $\theta_{vp,1}$, $\theta_{vp,2}$ are the vanishing angles of panorama 1 and panorama 2, and ${}^{2}\theta_{1}$ corresponds to the relative rotation of panorama 1's pose in the room Cartesian coordinate system of panorama 2, i.e. of ${}^2\textbf{T}_1$. 
 We then rotate panorama 1's room vertices (in panorama 2's frame) about the W/D/O midpoint, and recompute $\hat{\textbf{T}} = (\hat{x},\hat{y},\hat{\theta})$ by least-squares fitting between point sets to obtain $\hat{\textbf{T}}_{corrected} = (\hat{x}^\prime, \hat{y}^\prime, \hat{\theta}^\prime)$ with fixed wall thickness.

\subsection{Details on Global Pose Estimation}
\label{sec:pgo-spanning-tree-details}
\begin{figure}[h]
    \centering
    \includegraphics[width=0.8\columnwidth]{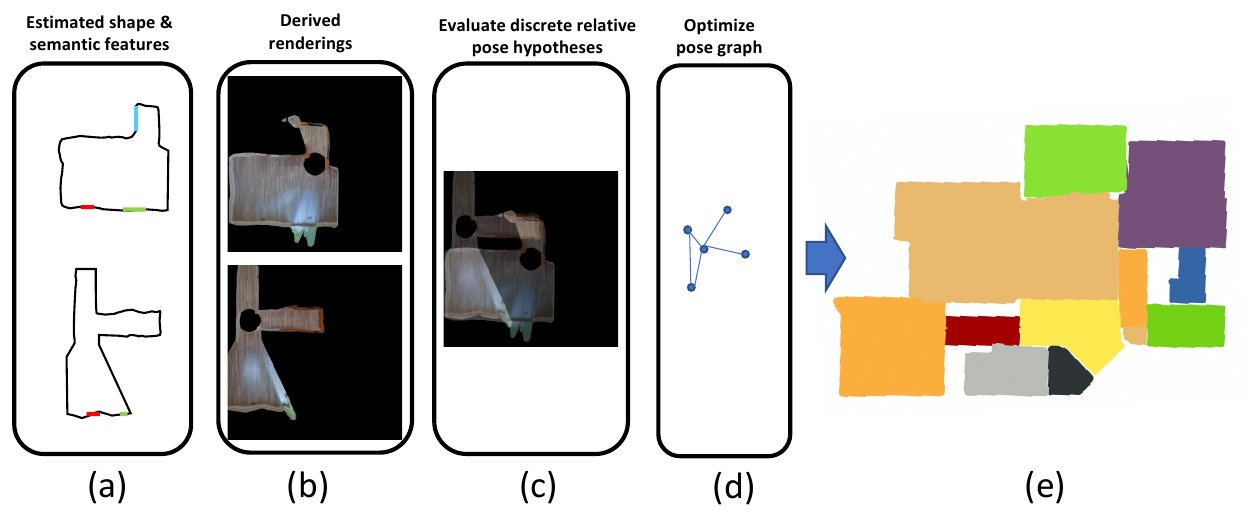}
    \caption{Starting with sparse panoramas (1-3 per room), in (a) we infer layout and semantic elements (\textbf{w}indows, \textbf{d}oors, \textbf{o}penings, or W/D/O). From these, in (b) we generate birds eye view (BEV) renderings of floors and ceilings (ceilings not shown here).  Next, plausible pairwise relative poses are hypothesized based on matching W/D/O. Each is accepted or rejected (c), by feeding the hypothesis-aligned renderings into our learned SALVe verifier.   This example shows two aligned renderings computed by hypothesizing that a window can be used to align both shapes  Brighter areas indicate overlap regions.   SALVe is trained to evaluate these aggregated overlap regions and output an accept/reject decision about whether the hypothesized relative pose is plausible. From the plausible relative poses, a pose graph is created and optimized (d). This allows room layouts to be positioned in a world coordinate system and fused into a final reconstructed raster floorplan (e). }
    \label{fig:another_overview}
\end{figure}

Here we provide details on the pose estimation and optimization referred to in Section \ref{sec:global-pose-estimation-optimization} of the main paper.

\noindent \textbf{No optimization (unfiltered spanning tree).} For $N$ images, the global motion can be parameterized by $N - 1$ motions. In the pose graph $\mathcal{G}=(\mathcal{V},\mathcal{E})$, when the graph G has a single connected component, the spanning tree is a set of edges such that every vertex in $\mathcal{V}$ is reachable from every other vertex in $\mathcal{V}$. For a graph with $N$ vertices, the minimum spanning tree always has $N - 1$ edges. However, global pose estimation with this method is inherently susceptible to error due to contamination by outliers.
To describe the method to compute a spanning tree, we assume the images are randomly ordered. Starting from the first image as the root, we incrementally include images in sequence, adding each next image into the current tree at its shortest path from the root.



\noindent \textbf{Pose Graph Optimization.}  Spanning tree solutions are susceptible to outlier edges in the tree. In order to exploit redundancy and utilize all the available information in the graph, {\em we use pose graph optimization}.


MAP inference for SLAM problems with Gaussian noise models is equivalent to solving a nonlinear least squares problem \cite{Dellaert21ar_FactorGraphs}. MAP inference comes down to maximizing the product of all factor graph potentials:

  \begin{equation}
    \mathbf{T}^{\textrm{MAP}} = \mbox{arg } \underset{T}{\mbox{max}} \prod_{(i,j) \in \mathcal{E}} \phi_{ij} (\mathbf{T}_i, \mathbf{T}_j) ,
\end{equation}
where $\phi_{ij}(\mathbf{T}_i, \mathbf{T}_j)$ is a factor graph potential:
\begin{equation}
    \phi_{ij}(\mathbf{T}_i, \mathbf{T}_j) \propto \mbox{exp} \Bigg\{ - \frac{1}{2} \| h_{ij}(\mathbf{T}_i, \mathbf{T}_j) - z_{ij} \|^2_{\Sigma_{ij}} \Bigg\} ,
\end{equation}
\noindent with $h_{ij}(T_i, T_j) = {T_i}^{-1}\cdot T_j$ and $z_{ij}$ is the estimated relative pose between images $i$ and $j$ from the alignment step described earlier.
 
The following objective function is then optimized using GTSAM:
\begin{equation}
  \mbox{arg } \underset{ \mathbf{T} }{\mbox{min}} \sum\limits_{(i,j) \in \mathcal{E}} \rho\Big( \|h_{ij}(\mathbf{T}_i, \mathbf{T}_j) - z_{ij} \|^2_{\Sigma_{ij}} \Big) ,
\end{equation}
\noindent making updates $\mathbf{T}_i \oplus \xi := \mathbf{T}_i \circ \mbox{exp}(\hat{\xi})$, where $\xi \in \mathfrak{se}(2)$.  
Here, $\rho(\cdot)$ is a Huber noise model.

MAP inference over the pose graph with Gaussian noise models \cite{Dellaert21ar_FactorGraphs} is done by maximizing the product of all factor graph potentials. We initialize the solution from a greedy spanning tree and then optimize using GTSAM \cite{Dellaert12_GTSAM}. We follow the official GTSAM PGO implementation example
\footnote{\url{https://github.com/borglab/gtsam/blob/develop/python/gtsam/examples/Pose2SLAMExample.py}}. Once the pose graph is optimized, we use the estimated poses and room layout to create the final floorplan.








\subsection{Ablation Experiments Using Oracle W/D/O Detection}
\label{sec:ablation-gt-wdo-gt-layout}
In this section, we perform ablation experiments comparing global pose estimation and floorplan reconstruction results with estimated W/D/O locations and estimated layout, vs. a baseline that has access to ground truth W/D/O detections and ground truth layout.

\noindent \textbf{How much worse is performance with predicted W/D/O and predicted layout vs. annotated D/W/O and annotated layout?} In Table~\ref{tab:global-pose-estimation-gt-wdo-ablation}, we compute an upper bound for the completeness of our method, by using human-annotated W/D/O and human-annotated layout as input to the system. This measures the ability of the CNN to reason about photometric signal in a less noisy setting (there is still noise from HoHoNet). Note that annotations are not perfect.

The results indicate the already-strong localization precision of our system, with roughly similar camera pose estimation errors (in rotation and translation). We provide no vanishing-point axis-alignment post-processing to these generated relative poses, which leaves the ground-truth (GT) based system susceptible to higher rotation errors. However, the translation error of the model with access to GT W/D/O is still lower on average (22 cm vs. 25 cm).

With GT layout and GT W/D/O, the floorplan IoU is 91\% higher -- 0.86 median IoU vs. 0.45 IoU with our predicted poses and layout. The percentage of cameras localized is also much higher (93.44\% vs. 57.14\%) than the system without access to GT. These results are extremely promising, underscoring the significant potential for further improving the floorplan reconstruction completeness of our system by improving the layout estimation and W/D/O detection network.

\begin{table*}[]
    \centering
    \caption{Ablation experiments on global pose estimation, comparing performance with estimated W/D/O locations and estimated layout, vs. performance with ground truth W/D/O locations and ground truth layout (\emph{oracle}).}
    \begin{adjustbox}{max width=1\columnwidth}
	\begingroup
    \begin{tabular}{l|cc|cc|cc| cc}
        \toprule
        \textsc{Input} & \multicolumn{2}{|c|}{\textsc{\textbf{Localization \%}}} & \multicolumn{2}{|c|}{\textsc{\textbf{Tour Avg. Rotation  }}}       & \multicolumn{2}{|c|}{\textsc{\textbf{Tour Avg. Translation }}} & \multicolumn{2}{|c}{\textsc{\textbf{Floorplan IoU }}} \\
                        & \multicolumn{2}{|c|}{}                                 & \multicolumn{2}{|c|}{\textsc{\textbf{Error (deg.)}}} & \multicolumn{2}{|c}{\textsc{\textbf{Error (meters)}}} & &  \\
                        &  \textsc{mean} & \textsc{median} & \textsc{mean} & \textsc{median}             & \textsc{mean} & \textsc{median} & \textsc{mean} & \textsc{median} \\
        \midrule
        \textsc{Predicted WDO + Predicted Layout} & 60.70 & 57.14 & \textbf{3.73} & \textbf{0.17} & \textbf{0.80} & 0.25 & 0.49 & 0.45   \\
        \textsc{GT WDO + GT Layout} & \textbf{88.58} & \textbf{93.44} & 5.02 & 0.21 & 0.98 & \textbf{0.22} & \textbf{0.78} & \textbf{0.86} \\
        \bottomrule
    \end{tabular}
    \endgroup
    \end{adjustbox}
    \label{tab:global-pose-estimation-gt-wdo-ablation}
\end{table*}

\subsection{Comparison with Extremal SfM \cite{Shabani21iccv_ExtremeSfM}}
\label{sec:eval-metrics-vs-shabani-extremal-sfm}

\begin{table*}[]
    \centering
    \caption{Summary of comparison of our method vs. that of \emph{Extremal SfM} by Shabani \emph{et al.} \cite{Shabani21iccv_ExtremeSfM}.}
    \begin{tabular}{l |c|c}
    \toprule
     & Extremal SfM \cite{Shabani21iccv_ExtremeSfM} & Ours \\
    \midrule
        Computational Complexity & Exponential Time $O(n!)$ & Polynomial Time $O(n^2k^2)$ \\
        \# Panoramas / Room & 1 & $\geq 1$ \\
        \# Panoramas / Floor & 3.4 & 23.2 \\
        \# Floors in Test Set & 46 & 291 \\
        Door Configuration & Opposite facing surface normals & Any configuration \\
        Supported Room Type & Small size, Little self-occlusion & Any \\
        Home Type & Apartment & Residential Home \\
        Wall Assumption & Manhattan & None \\
        Evaluated Error Types & Translation & Rotation and translation \\
        Input Signal & BEV mask & BEV (image) \\
        Verifier Type & GNN on tree-structured graph & CNN on pairwise renderings \\
        \bottomrule
    \end{tabular}
    \label{tab:shabani-comparison-brief}
\end{table*}

\begin{table*}
\centering
\caption{A more detailed comparison of our input, method, and evaluation vs. those of Shabani \emph{et al.} \cite{Shabani21iccv_ExtremeSfM}. }
\begin{adjustbox}{max width=\columnwidth}
\begingroup
\noindent \begin{tabular}{p{0.25\columnwidth}|p{0.875\columnwidth}|p{0.875\columnwidth} }
\toprule
 \textsc{Comparison Type}    & \textsc{Extremal SfM (Shabani et al. \cite{Shabani21iccv_ExtremeSfM} } & \textsc{Ours} \\
 \midrule
\textsc{Input} & \begin{itemize}
  \item Assumes no room shape overlap in dataset. \vspace{1mm}
  \item Requires exactly one panorama per room. 
  \item Requires each input panorama to see most of the room, including W/D. This would be problematic for complex rooms where one panorama sees only a fraction of the room. 
  \item Demonstrated on apartments.
\end{itemize}
& \begin{itemize}
  \item Handles any amount of overlap, but not zero overlap. 
  \item Requires one or more panoramas per room. 
  \item Has no requirement on panorama capture locations. 
  \item Evaluated on ZInD with complex room layouts, including open floorplans.
  \end{itemize} \\
  \midrule
  \textsc{Method} & 
  \begin{itemize}
  \item Uses W/D alignment, but not openings. \vspace{1mm}
  \item Uses HorizonNet \cite{Sun19cvpr_HorizonNet} for layout and separately predicted W/D objects at test time. 
  \item Uses room topologic information and BEV semantic masks (with no photometric information) directly as input to GNN. 
  \item Relies on a tree-type graph topology. They verify on all possible global graph configurations (room snapping combinations) with graph neural network Conv. MPN \cite{Zhang20cvpr_ConvMPN}.
  \item Runs in exponential time with number of rooms.
  \end{itemize}
  & 
   \begin{itemize}
  \item Uses W/D/O alignment to generate pairwise initial pose hypotheses. 
  \item Uses predicted room layout from HorizonNet \cite{Sun19cvpr_HorizonNet} with joint W/D/O predictions and wall-floor boundary uncertainty. 
  \item Uses BEV photometric signal in panorama pairwise pose verification. 
  \item Uses global filtering and optimization similar to traditional Global SfM. Our method is also able to refine coarse panorama poses. 
  \item Runs in polynomial time with number of rooms. 
  \end{itemize}
  \\
  \midrule \\
  \textsc{Evaluation} & 
   \begin{itemize} 
  \item Evaluated on a small dataset of 46 apartments. The dataset contains 3.4 panoramas per apartment with all Manhattan room layouts. \vspace{1mm}
  \item Generates top-5 possible floorplans. 
  \item Localization is considered a success if any of K possible solutions has estimated global poses with mean positional error below $\delta$ meters. Rotation error is not considered. 
  \end{itemize}
  & 
   \begin{itemize}
  \item Evaluated on the test split of ZInD with 291 floorplans of residential homes. ZInD contains 23.2 panoramas per floorplan with Manhattan and non-Manhattan room layouts. 
  \item Generates 1 final floorplan.
  \item Evaluated by per-panorama average pose rotation and translation error, and localization completeness.
  \end{itemize}
  \\
  \bottomrule
\end{tabular}
\endgroup
\end{adjustbox}
\label{tab:shabani-comparison-extended}
\end{table*}

In this section, we provide additional comparisons with concurrent work by Shabani \emph{et al.} \cite{Shabani21iccv_ExtremeSfM} (See Tables \ref{tab:shabani-comparison-brief} and \ref{tab:shabani-comparison-extended}).

\noindent \textbf{Differences in Assumptions.}
\begin{itemize}
    \item Shabani \emph{et al.}'s one-panorama-per-room assumption limits the number of door hypotheses, as they assume door surface normals must point in opposite directions, whereas we consider twice as many hypotheses, i.e. when surface normals may also point in the same direction. The restricted door hypotheses would be analogous to querying a ZInD oracle for ground truth adjacency, which we do not do.
\end{itemize}





\noindent \textbf{Differences in Method.}

\begin{itemize}
    \item They do not use openings. Accordingly, the rooms cannot be too large or complex enough to have significant amounts of self-occlusion.
    \item Exponential time: They rely upon a tree-type graph topology. They verify on all possible global graph configurations (room snapping combinations) with graph neural network. Conv MPN \cite{Zhang20cvpr_ConvMPN} machinery (follows up on HouseGAN \cite{Nauata20eccv_HouseGAN} and HouseGAN++ \cite{Nauata21cvpr_HouseGAN++} machinery). This requires exponential time.
    
\end{itemize}


\noindent \textbf{Differences in Evaluation.}
\begin{itemize}
    \item Instead of computing a mean error per panorama, they just count the successes. This does not take into account catastrophic failures (see their Figure 9).  We have been evaluating as ``you get one shot, and for every pano you try to localize, it will go into your mean error''. By comparison, they find the minimum subset out of all panoramas they localized that are good.
\end{itemize}



Unfortunately, at the time of our submission their code and dataset used were not publicly available, so a comparison of accuracies on a common dataset is not possible at this time.




\subsection{Analysis of Computational Complexity}
\label{sec:salve-computational-complexity}
We compute over the ZInD train/val/test sets putative estimates of these constants $n,k$. On average for each ZInD tour, we find $n \approx 23.2$. When evaluating  $\mathcal{O}(n^2k^2)$, we find that each floor has on average $10795$ (mean) and $8188.0$ (median) putative alignments.

After pruning away impossible hypotheses via width ratios,
$n$ is unchanged but $k$ is reduced, leaving  $\mathcal{O}(n^2k^2) \approx 5804.5$ (mean) and $3441.0$ (median). Although there are just as many panoramas to match with, we effectively have fewer instances of each W/D/O type we can feasibly match with (thus reducing $k$).


This leads to a highly imbalanced classification problem, with negative-to-positive ratios of 18:1 on average, when using predicted layout and predicted W/D/O locations.  



\noindent \textbf{Oracle Layout Generator Baseline.} For alignments generated from GT W/D/O and GT layout, we halve the computational complexity by discarding those with penetrated freespace (an average negative-to-positive ratio of 7:1). For predicted layout, we cannot prune alignments by freespace penetration heuristics due to arbitrary predicted locations of openings in large rooms.



\subsection{Details on Layout-Only Rasterization Baseline}
\label{sec:layout-only-visual-examples}
\begin{figure}
    \centering
    \subfloat[Positive example pair 1.]{ 
        \includegraphics[width=0.2\columnwidth]{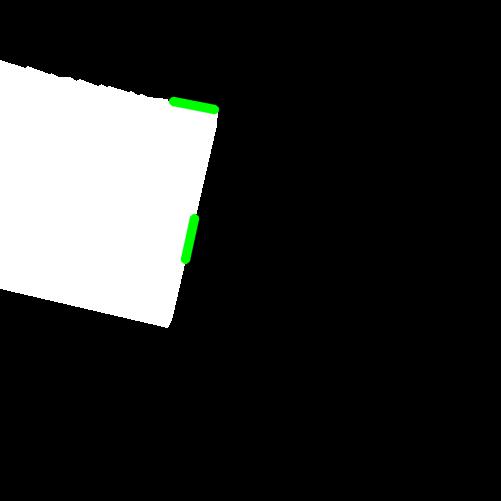}
        \includegraphics[width=0.2\columnwidth]{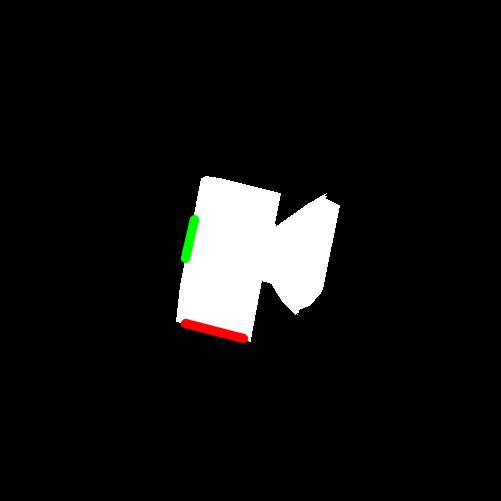}
        \includegraphics[width=0.2\columnwidth]{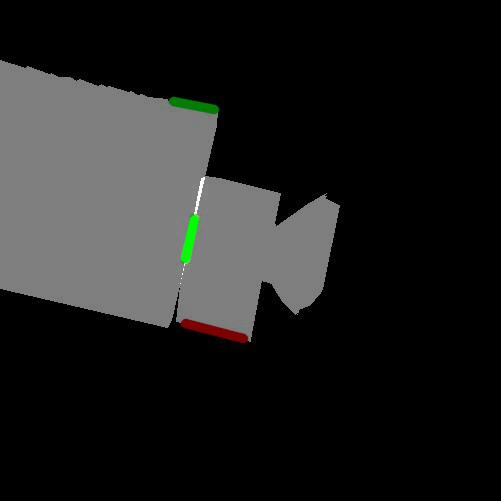}
    }
    \vspace{2mm}
    \subfloat[Positive example pair 2.]{ 
        \includegraphics[width=0.2\columnwidth]{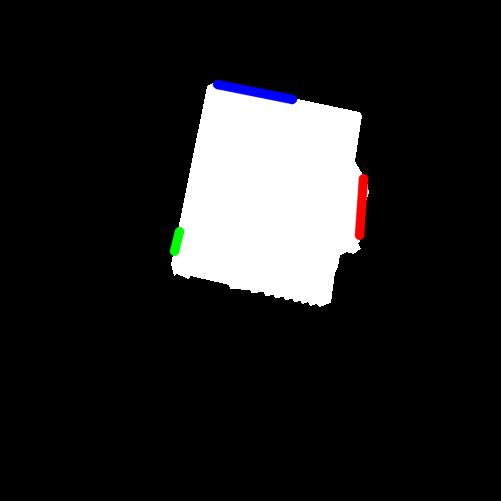}
        \includegraphics[width=0.2\columnwidth]{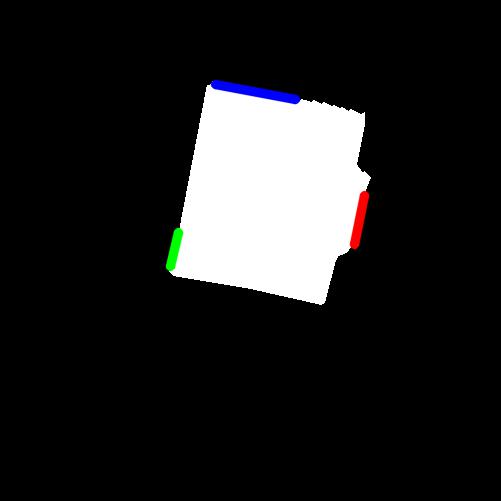}
        \includegraphics[width=0.2\columnwidth]{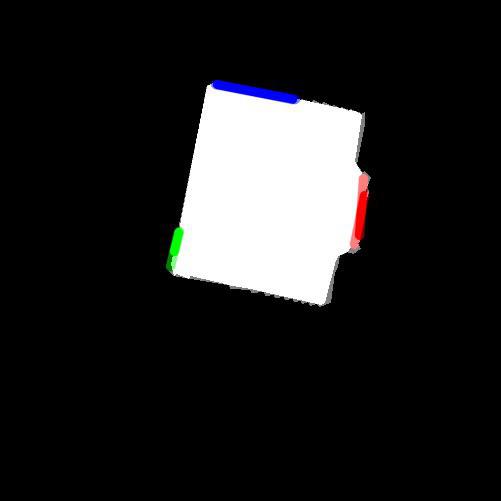}
    }
    \vspace{2mm}
    \subfloat[Negative example pair 1.]{ 
        \includegraphics[width=0.2\columnwidth]{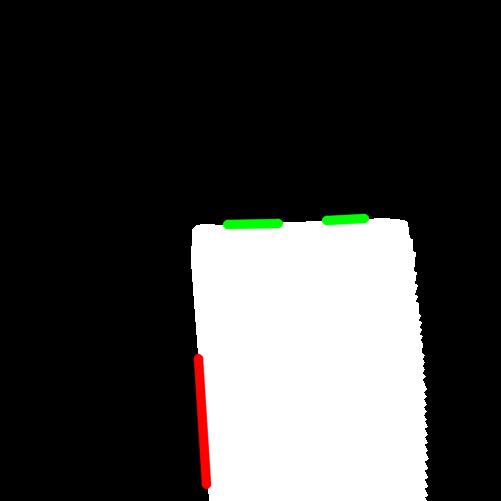}
        \includegraphics[width=0.2\columnwidth]{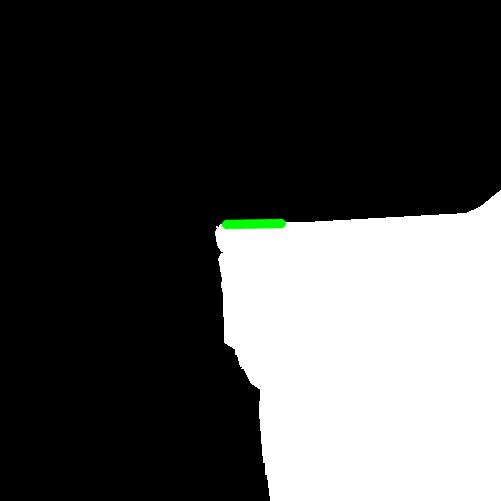}
        \includegraphics[width=0.2\columnwidth]{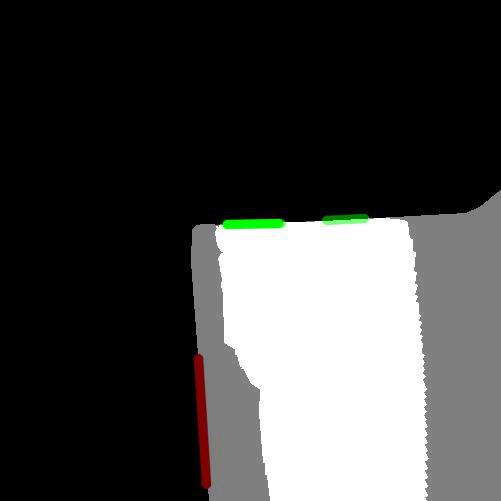}
    }
    \vspace{2mm}
    \subfloat[Negative example pair 2.]{ 
        \includegraphics[width=0.2\columnwidth]{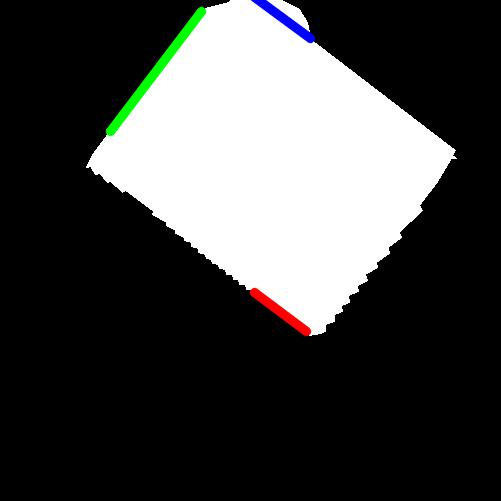}
        \includegraphics[width=0.2\columnwidth]{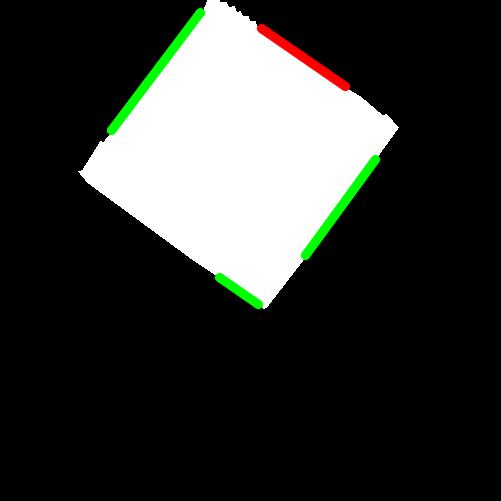}
        \includegraphics[width=0.2\columnwidth]{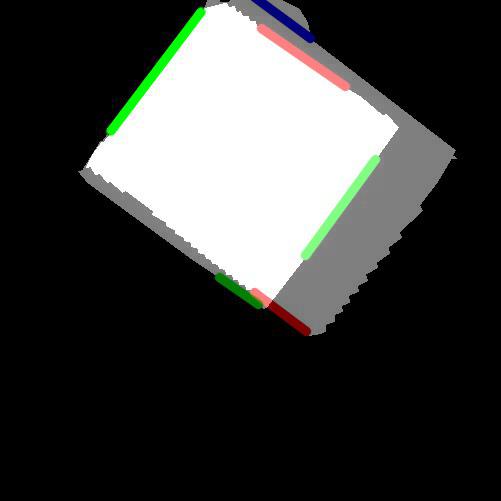}
    }
    \caption{Examples of layout-only rasterized input. Each row represents an alignment pair.  \textbf{\emph{Left:}} rendering for panorama 1. \textbf{\emph{Middle:}} rendering for panorama 2. \textbf{\emph{Right:}} blended images (for visualization only).}
    \label{fig:layout-only-input-examples}
\end{figure}

In Table \ref{tab:global-pose-estimation-ablation} (Section 6) of the main paper, we 
reported results of a model that has no access to photometric information as input, but only to rasterized BEV layout. In Figure \ref{fig:layout-only-input-examples}, we show examples of such input.

\subsection{Additional Discussion Points}
\label{sec:salve-misc-details-discussion-intuition}

\subsubsection{Accuracy vs. Amount of Visual Overlap}

\begin{figure}
    \centering
    \subfloat[Classification Accuracy vs. Visual Overlap (for GT Positives vs. Negatives)]{
        \includegraphics[width=0.3\columnwidth]{latex/figs/accuracy_vs_overlap/2021_10_26__ResNet152__435tours_serialized_edge_classifications_test2021_11_02___bar_chart_acc_positives_only__confthresh0.0.pdf}
        \includegraphics[width=0.3\columnwidth]{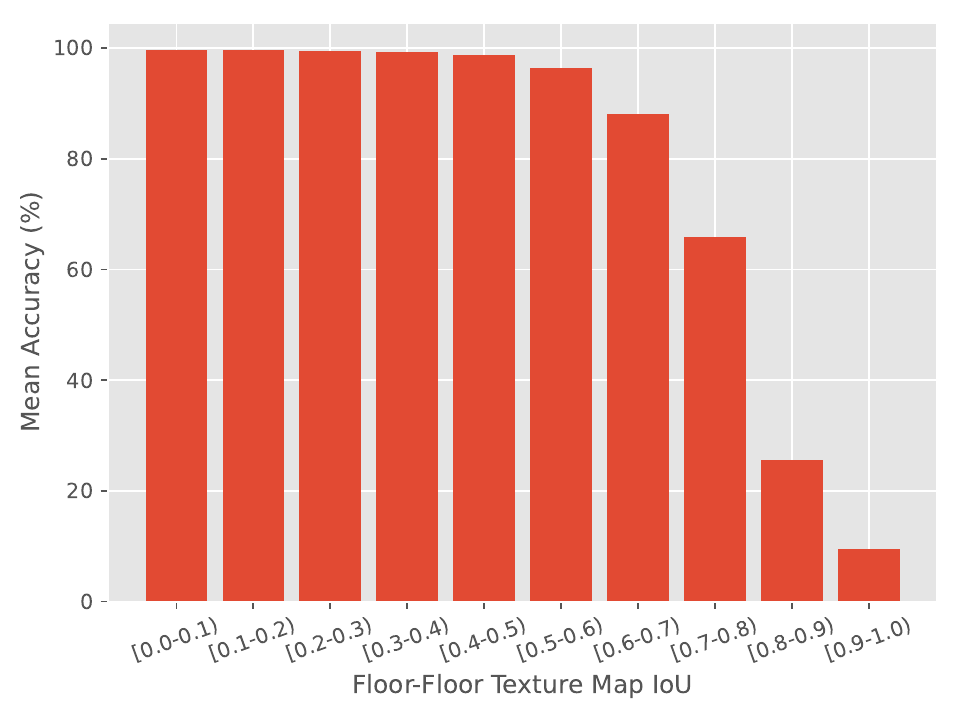}
    }
    \hspace{2mm}
    \subfloat[Rotation and Translation Error vs. Visual Overlap (for GT Positives )]{
        \includegraphics[width=0.3\columnwidth]{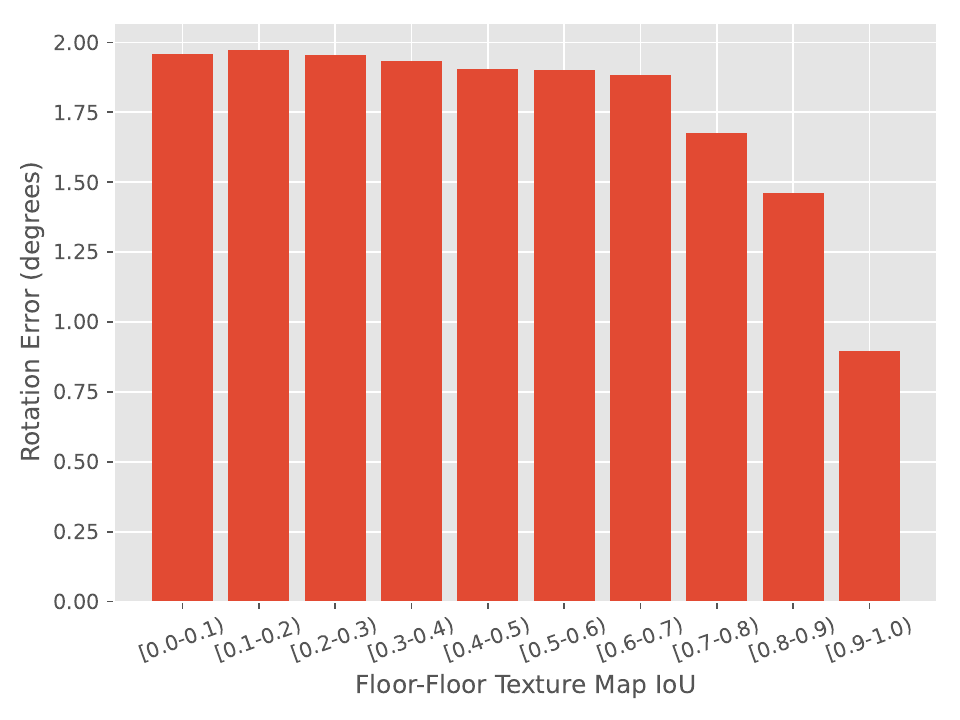}
        \includegraphics[width=0.3\columnwidth]{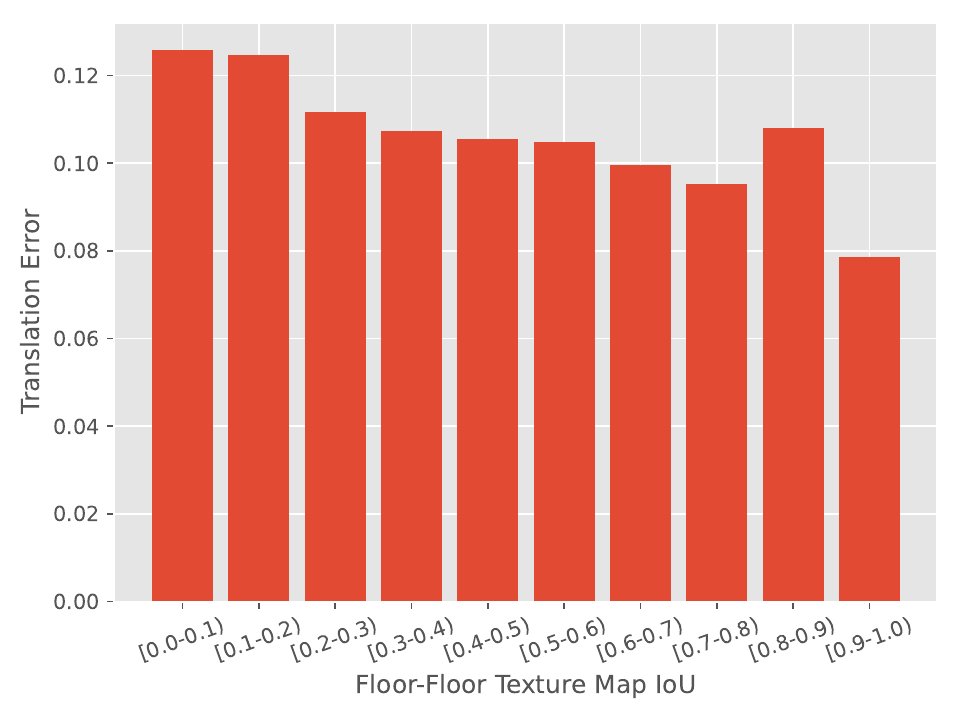}
    }
    \hspace{2mm}
    \subfloat[Rotation and Translation Error vs. Visual Overlap (for GT Negatives )]{
        \includegraphics[width=0.3\columnwidth]{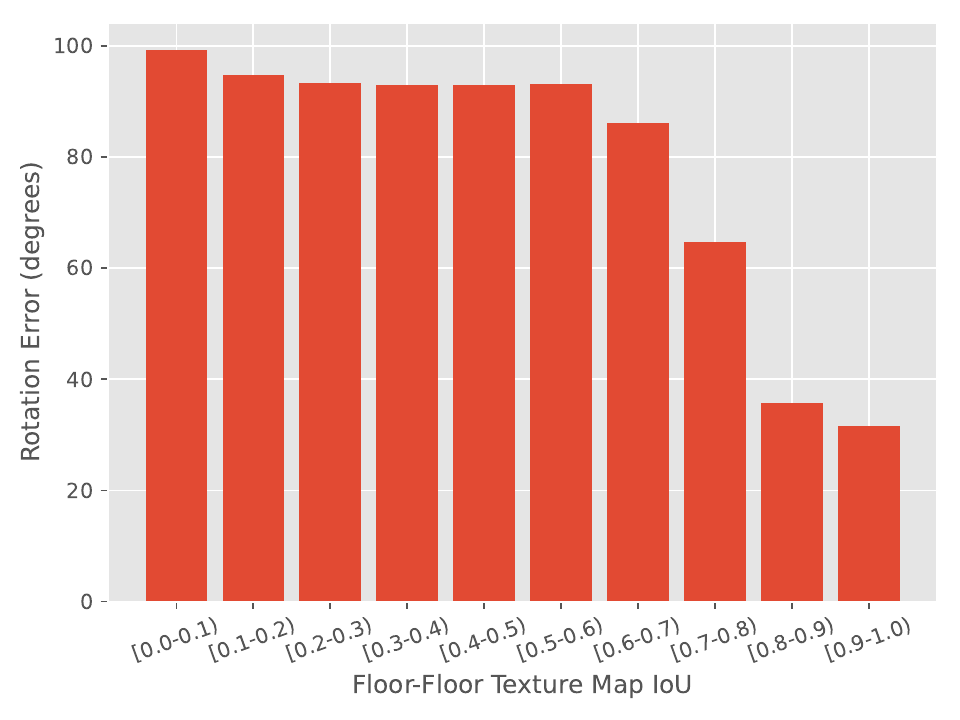}
        \includegraphics[width=0.3\columnwidth]{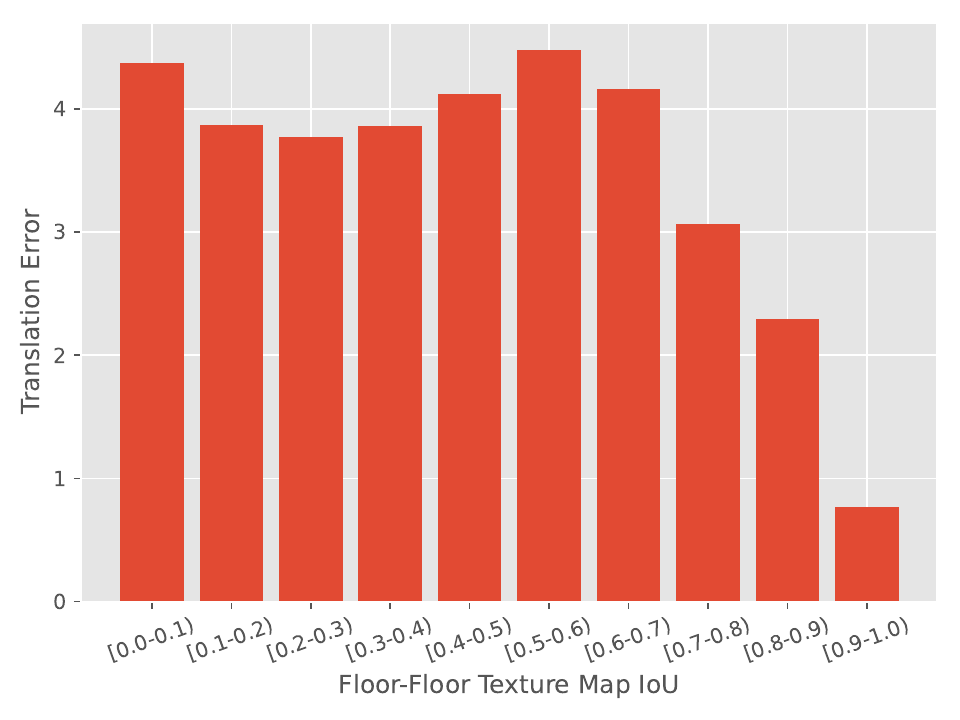}
    }
    \hspace{2mm}
    \subfloat[Visual Overlap Distribution in ZinD (for GT Positives vs. Negatives)]{
        \includegraphics[width=0.3\columnwidth]{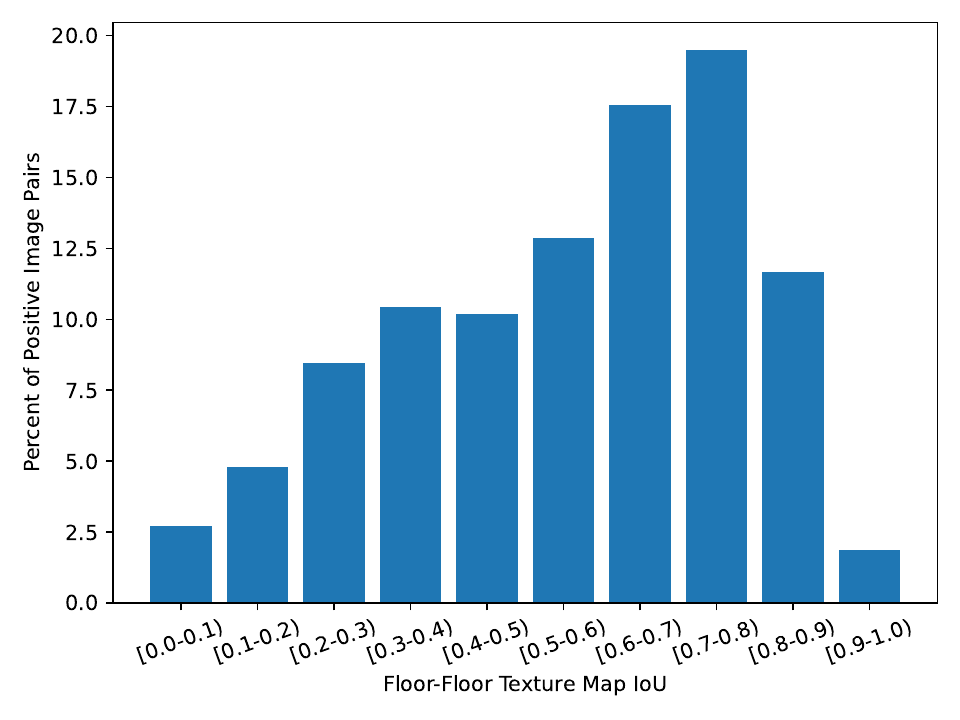}
        \includegraphics[width=0.3\columnwidth]{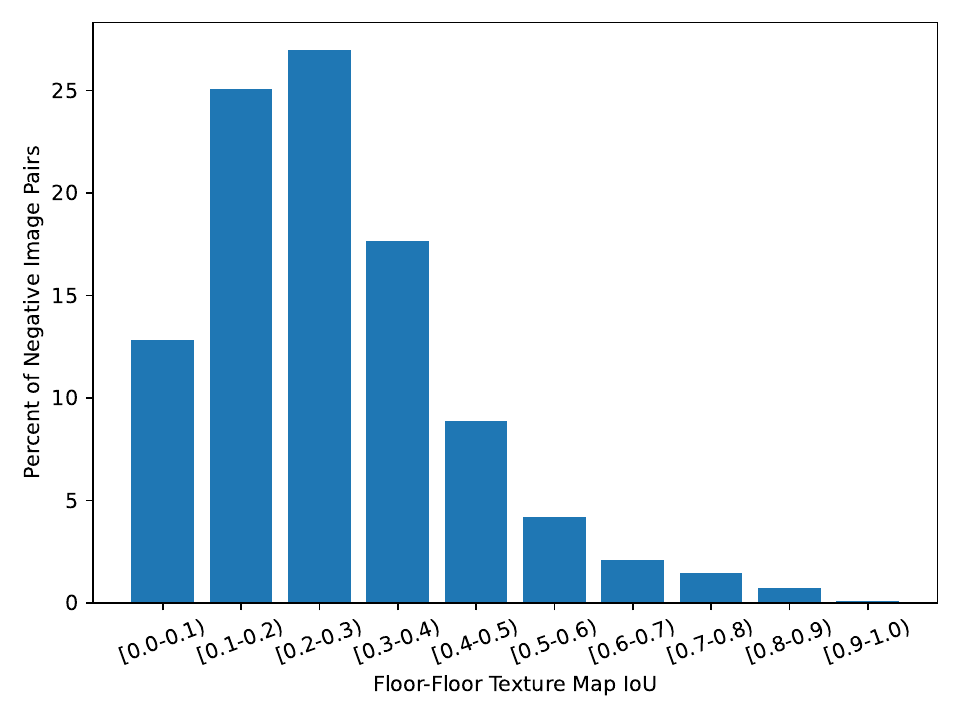}
    }
    \caption{ \textbf{(Row 1)} Classification accuracy vs. overlap for the GT \textbf{positive} class only \textbf{(left)} and \textbf{negative} class only \textbf{(right)} for ResNet-152 model.   \textbf{(Row 2)} Relative pose rotation error \textbf{(left)} and translation \textbf{(right)} vs. amount of visual overlap for GT \emph{positive} examples. \textbf{(Row 3)} Relative pose rotation error \textbf{(left)} and translation \textbf{(right)} vs. amount of visual overlap for GT \emph{negative} examples.   \textbf{(Row 4)} Distribution of visual overlap (IoU) over rendered buildings for positive pairs \textbf{(left)} and negative pairs \textbf{(right)} from SE(2) alignments generated from predicted W/D/O's. }
    \label{fig:accuracy-vs-overlap-additional-analysis}
\end{figure}

Poor accuracy for examples with small support in the dataset (see Figure \ref{fig:accuracy-vs-overlap-additional-analysis}) shows the potential for hard negative mining in future work. Negative examples with high IoU (see Figure \ref{fig:accuracy-vs-overlap-additional-analysis}d, right) are few in the dataset, and present high error (see Figure \ref{fig:accuracy-vs-overlap-additional-analysis}a, right).

\subsubsection{Additional Details on Evaluation Metrics}

To compute both the ground truth mask and estimated binary mask for IoU computation, we rasterize the scene to a grid resolution of $10 \times 10$ centimeters per cell (we found an even finer resolution did not affect results significantly). \\

\noindent \textbf{Error Rate in Supervision Generation.} Noisy generation of ground truth is susceptible to false negatives. In other words, the generator falsely assumes that there are no `positive' alignments for certain panorama pairs, due to errors just above the maximum tolerated rotation or translation thresholds. At inference time, these lead to false positive predictions, as the model identifies these pairs as positives, conflicting with the ground truth. We find that for  spatially adjacent rooms, no putative `positive' hypothesis is generated for less than 9.67\% of panorama pairs, due to layout estimation or W/D/O prediction errors. 


\subsubsection{Model Learning}

Previous work has proven that domain-knowledge of indoor space, such as room intersections, loop closure, and multi-view alignment, can be helpful in solving the `room merge' problem \cite{Cruz21cvpr_ZillowIndoorDataset}. On the other hand, visual overlap of floor and ceiling areas from different texture images provides helpful clues, such as light source reflections, paneling direction of wood flooring, and shared ceiling features, to verify panorama registration \cite{Dellaert1999cvpr_CondensationAlgorithmLocalization}. We extract undistorted floor and ceiling orthographic views from each panorama using inferred depth and register each view using an estimated W/D/O alignment hypothesis. While inferred depth signal alone suffers from inaccuracies at very close or far range and near reflective objects such as mirrors, the orthographic views still contain small distortion and therefore provides a strong signal for alignment verification. We train a model to implicitly verify the aligned texture signals (such as light source reflections, paneling direction of wood flooring, and shared ceiling features), as well as model other priors on room adjacency, such as the fact that bathrooms and bedrooms are often adjacent.

While ceiling features on a mosaiced ceiling image have been used for robot localization for at least two decades \cite{Dellaert1999cvpr_CondensationAlgorithmLocalization}, success of registration using traditional explicit image-based matching is highly dependent on significant appearance similarity. This is typically not the case for our work, due to the large baselines and potentially very different times of capture.

Aligned orthographic views can help identify shared floor texture around room openings, identifying common objects (i.e. refrigerators), or known priors on room adjacency, such as the fact that bathrooms and bedrooms are often adjacent. axis-alignment of walls between two layouts. Because floorplan adjacency is governed by strong priors, such as primary bedrooms are attached to primary bathrooms, such signals can be learned.

Whereas previous works have employed domain-knowledge to manually define features for room-merge costs to employ in ranking \cite{Cruz21cvpr_ZillowIndoorDataset}, we set out to learn such features from data. Previous work has defined costs that aim to minimize room intersections, maximize loop closure, maximize multi-view alignment of semantic elements, and produce the most axis-aligned floorplans \cite{Cruz21cvpr_ZillowIndoorDataset}. However, some of these costs are only applicable if an annotator can identify which panoramas were captured in different rooms, as layouts within the same room should instead \emph{maximize} room intersection, while those captured in separate rooms should \emph{minimize} room intersection. Each window from panorama $\mathbf{I}_a$ should reproject onto a window in panorama $\mathbf{I}_b$ only if the panoramas were captured in the same room, which is an unknown latent variable. IoU should be high between the two overlaid room-layouts; however, this is not true if they are in separate rooms (cross-room). If, on the other hand, we knew a ``same-room'' label, then layout-IoU would be useful during localization.

\noindent \textbf{Available Signals for Learning Priors.} Many possible complementary signals can be employed in the reconstruction problem. The $360^\circ$ image suffers from significant distortion, while inferred depth suffers from inaccuracies at very close or far range and near reflective objects such as mirrors. Taken together, however, undistorted texture can be extracted in an orthographic manner. As the floor alone can have highly homogeneous texture or varied lighting, the ceiling can also provide helpful clues. Registration of inferred layout alone, with consideration of the image content, can lead to implausible arrangements.  No single signal is sufficient.

Human annotators use a variety of different cues to solve the merge task, most of them grounded in visual features within the image. For example, they often rely upon identifying shared floor texture around room openings, identifying common objects (i.e. refrigerators), or known priors on room adjacency, such as the fact that bathrooms and bedrooms are often adjacent. axis-alignment of walls between two layouts. Because floorplan adjacency is governed by strong priors, such as primary bedrooms are attached to primary bathrooms, such signals can be learned. Additional such relationships include hallway-to-bedroom adjacency.

\subsection{Additional Examples of Illumination Changes}
\label{sec:extreme-illumination-change-example}

\begin{figure*}
    \centering
    \subfloat[Input panorama pair.]{
    
        \includegraphics[width=0.5\columnwidth]{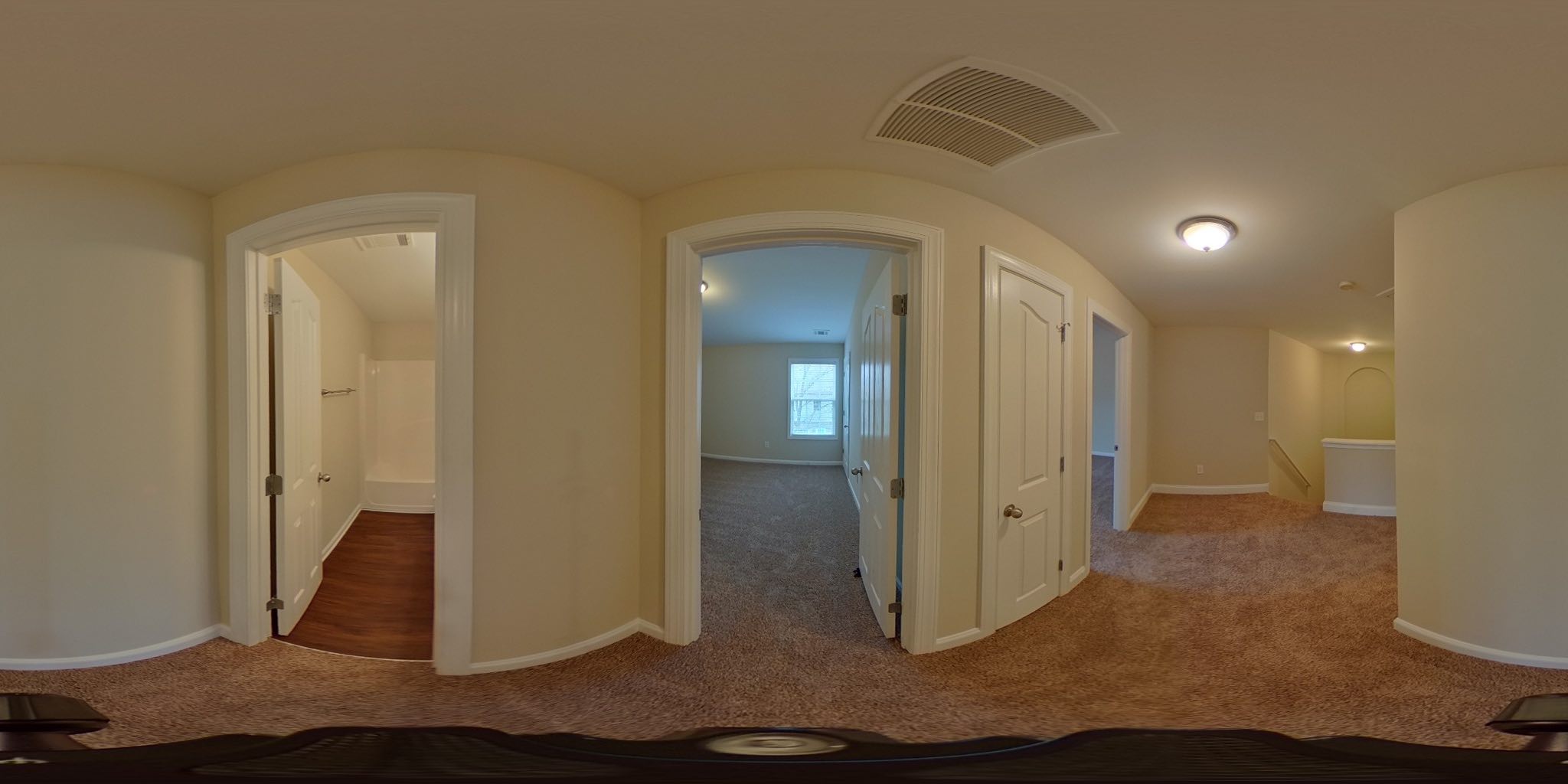}
        \includegraphics[width=0.5\columnwidth]{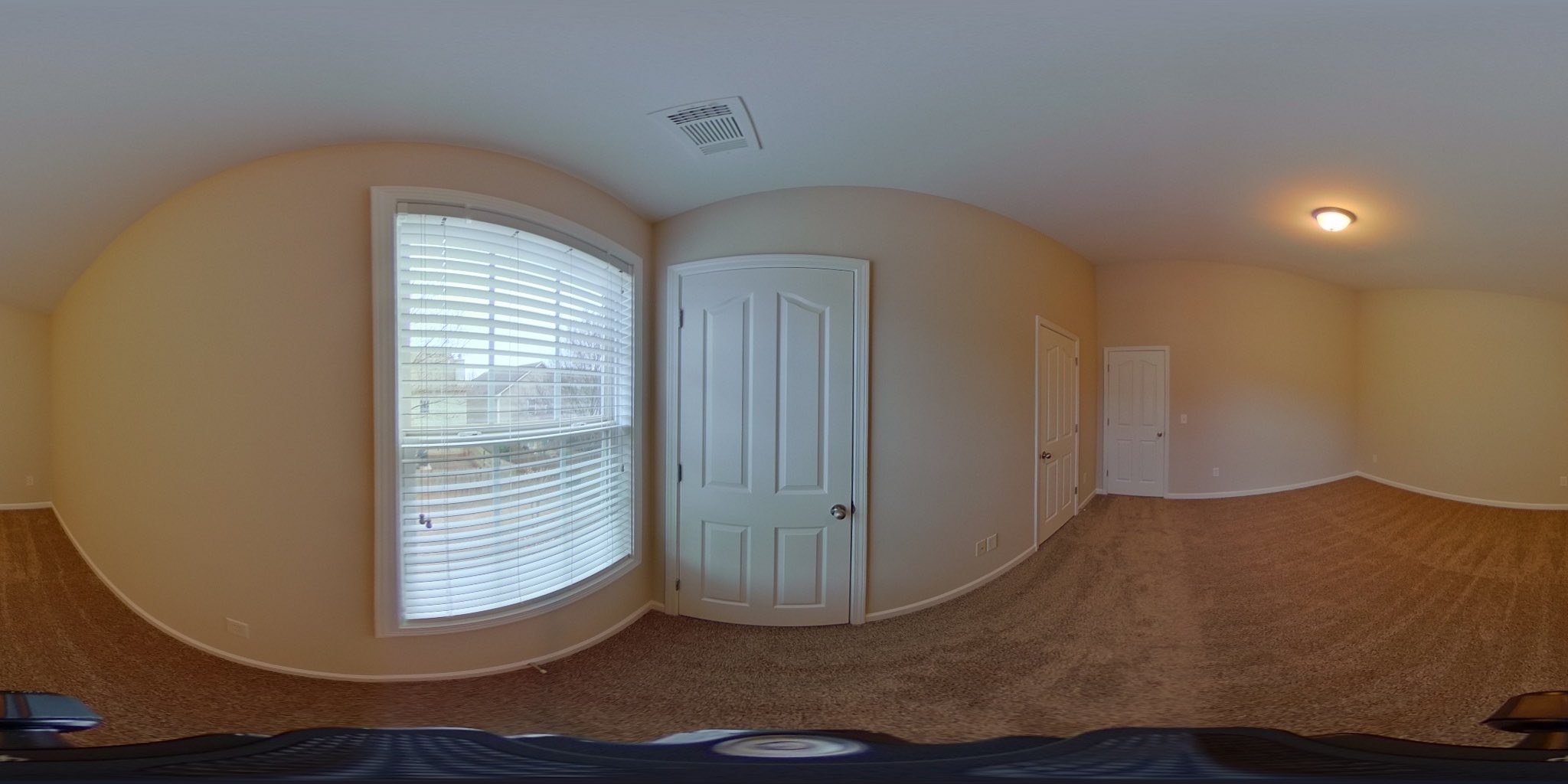}
    }
    \vspace{2mm}
    \subfloat[Orthographic floor texture maps, with an extreme illumination change.]{
        \includegraphics[width=0.33\columnwidth]{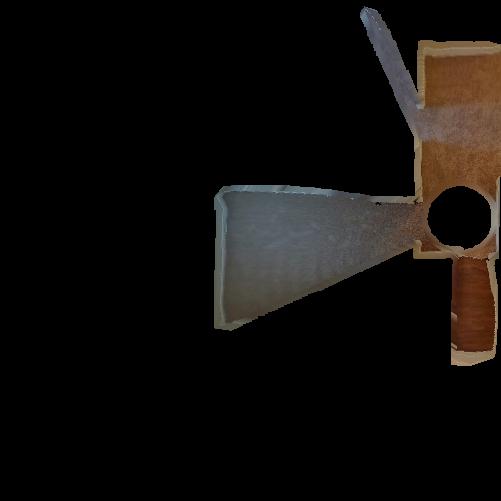}
        \includegraphics[width=0.33\columnwidth]{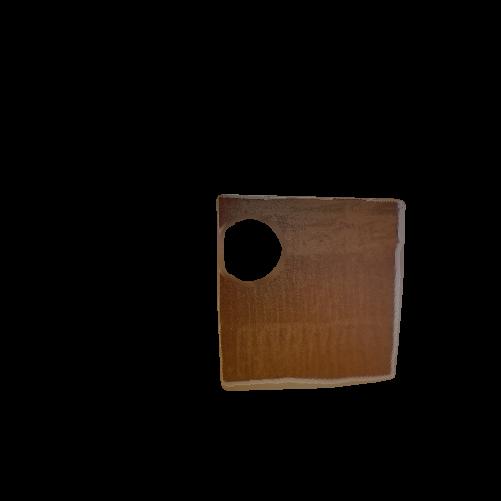}
        \includegraphics[width=0.33\columnwidth]{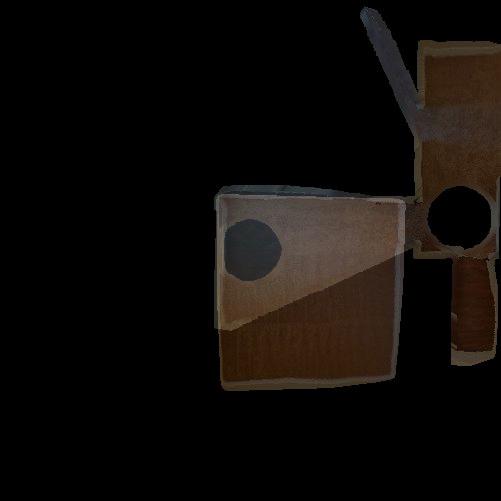}
    }
    \vspace{2mm}
    \subfloat[Orthographic ceiling texture maps.]{
        \includegraphics[width=0.33\columnwidth]{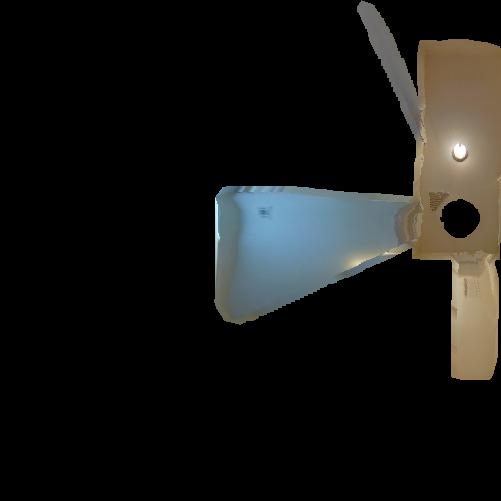}
        \includegraphics[width=0.33\columnwidth]{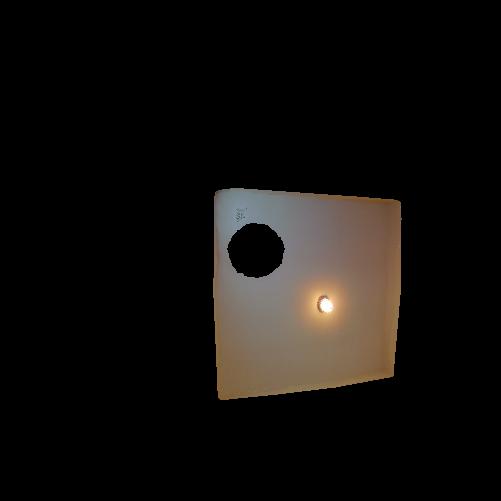}
        \includegraphics[width=0.33\columnwidth]{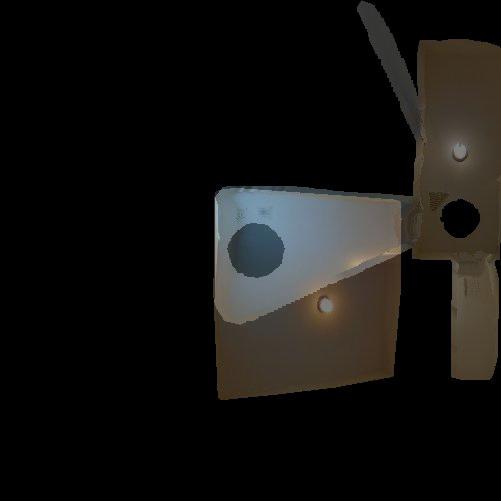}
    }
    \caption{Example of an extreme illumination change, as the carpet color appears to shift from brown to grey \textbf{(middle)}, and ceiling from warm yellow to light blue \textbf{(bottom)}.}
    \label{fig:extreme-illumination-change}
\end{figure*}

In Figure \ref{fig:extreme-illumination-change}, we provide an example of extreme illumination changes in ZInD, which prevent the use of classical image alignment algorithms for BEV image registration.

\subsection{Verifier Data Augmentation and Training Details}

\noindent \textbf{Verifier data augmentation.} We resize BEV texture maps to $234 \times 234$ resolution, sample random $224 \times 224$ crops, randomly flip them, and then normalize crops using the ImageNet mean and standard deviation. 

\noindent \textbf{Verifier training.} We use a ResNet-152 architecture with ImageNet-pretrained weights, training for 50 epochs, with an initial learning rate of $1 \times 10^{-3}$, polynomial learning rate decay with a decay factor of $0.9$ per iteration, and a weight decay of  $1 \times 10^{-4}$. We use a batch size of 256 examples on 3 NVIDIA Quadro RTX 6000 GPUs.

\subsection{Ethical/Privacy/Transparency/Fairness/Social Impact Concerns}
\label{sec:ethical-privacy-concerns}
Floor plan reconstruction, in general, could potentially lead to privacy issues. However, schematic floorplans are able to abstract away details of the real interior space, thereby revealing the layout and functionality of a home while hiding personal information (PI) and personally identifiable information (PII) information from the images used to reconstruct it. In other words, we can build the floorplan from 360$^\circ$ panos and then immediately use the floorplan as a medium (to convey the space), while suffering from fewer privacy issues compared to releasing all the 360$^\circ$ images used to create it.

\end{document}